\documentclass{article}


\usepackage[linesnumbered,ruled,vlined]{algorithm2e}
\usepackage{microtype}
\usepackage{graphicx}
\usepackage{subcaption}
\usepackage{booktabs} 
\usepackage{enumitem}
\usepackage{xspace}
\usepackage[dvipsnames]{xcolor}
\usepackage{multirow}
\usepackage{makecell}
\usepackage{esvect}
\usepackage{xurl}
\usepackage{hyperref}
\usepackage{hyperref}
\usepackage{float}
\usepackage{wrapfig}

\newcommand{\ksweepmedium}{
\begin{table}[H]
\centering
\small
\begin{tabular}{lccccc}
\toprule
\textbf{Environment} & \textbf{k=5} & \textbf{k=10} & \textbf{k=25} & \textbf{k=50} & \textbf{k=75} \\
\midrule
pointmaze-medium & $76.50 \pm 10.8$ & $\mathbf{80.75 \pm 3.2}$ & $\mathbf{84.00 \pm 0.7}$ & $\mathbf{81.00 \pm 3.4}$ & $72.75 \pm 4.9$ \\
pointmaze-large & $45.75 \pm 17.7$ & $61.00 \pm 13.2$ & $70.00 \pm 6.8$ & $\mathbf{88.25 \pm 5.5}$ & $\mathbf{88.75 \pm 4.8}$ \\
pointmaze-giant & $22.50 \pm 15.4$ & $47.50 \pm 14.1$ & $62.50 \pm 13.3$ & $67.75 \pm 12.4$ & $\mathbf{72.25 \pm 9.5}$ \\
pointmaze-teleport & $24.00 \pm 8.5$ & $26.50 \pm 4.0$ & $28.50 \pm 7.8$ & $27.75 \pm 6.8$ & $\mathbf{36.25 \pm 6.4}$ \\
antmaze-medium & $89.75 \pm 0.8$ & $\mathbf{95.75 \pm 1.4}$ & $\mathbf{96.50 \pm 1.1}$ & $90.75 \pm 2.5$ & $90.75 \pm 3.4$ \\
antmaze-large & $51.75 \pm 3.2$ & $74.75 \pm 1.4$ & $\mathbf{86.50 \pm 6.5}$ & $\mathbf{86.00 \pm 3.5}$ & $68.25 \pm 1.7$ \\
antmaze-giant & $3.75 \pm 2.1$ & $38.75 \pm 6.6$ & $\mathbf{73.00 \pm 2.5}$ & $57.75 \pm 4.7$ & $26.25 \pm 6.4$ \\
antmaze-teleport & $36.25 \pm 4.5$ & $41.50 \pm 6.1$ & $40.75 \pm 2.7$ & $\mathbf{48.00 \pm 3.1}$ & $43.25 \pm 5.8$ \\
antsoccer-medium & $6.25 \pm 2.8$ & $4.75 \pm 2.5$ & $\mathbf{12.25 \pm 1.7}$ & $\mathbf{12.25 \pm 1.1}$ & $6.50 \pm 0.5$ \\
antsoccer-arena & $\mathbf{58.75 \pm 6.2}$ & $53.25 \pm 3.6$ & $\mathbf{58.75 \pm 5.6}$ & $\mathbf{60.00 \pm 6.3}$ & $\mathbf{59.00 \pm 4.0}$ \\
\bottomrule
\end{tabular}
\vspace{3mm}
\caption{Sensitivity to horizon length $k$ on locomotion environments. Success rates (\%) averaged over 8 seeds with 95\% confidence intervals reported. Results within 95\% of the best are \textbf{bolded}.}
\label{tab:k_sweep_locomotion}
\end{table}
}

\newcommand{\ksweeplarge}{
\begin{table}[H]
\centering
\small
\begin{tabular}{lccccc}
\toprule
\textbf{Environment} & \textbf{k=25} & \textbf{k=75} & \textbf{k=100} & \textbf{k=125} & \textbf{k=150} \\
\midrule
humanoidmaze-medium & $\mathbf{88.00 \pm 4.4}$ & $\mathbf{85.00 \pm 1.7}$ & $\mathbf{87.00 \pm 3.0}$ & $\mathbf{86.67 \pm 5.4}$ & $\mathbf{87.33 \pm 4.6}$ \\
humanoidmaze-large & $44.67 \pm 5.6$ & $\mathbf{48.67 \pm 2.7}$ & $\mathbf{48.50 \pm 4.8}$ & $\mathbf{49.33 \pm 6.2}$ & $\mathbf{50.00 \pm 5.2}$ \\
humanoidmaze-giant & $18.00 \pm 4.7$ & $\mathbf{26.67 \pm 2.4}$ & $22.75 \pm 3.9$ & $\mathbf{28.00 \pm 5.9}$ & $21.67 \pm 0.5$ \\
\bottomrule
\end{tabular}
\vspace{3mm}
\caption{Sensitivity to horizon length $k$ on humanoidmaze environments. Success rates (\%) averaged over 8 seeds with 95\% confidence intervals reported. Results within 95\% of the best are \textbf{bolded}.}
\label{tab:k_sweep_humanoid}
\end{table}
}

\newcommand{\ksweepsmall}{
\begin{table}[H]
\centering
\small
\begin{tabular}{lccccc}
\toprule
\textbf{Environment} & \textbf{k=3} & \textbf{k=5} & \textbf{k=10} & \textbf{k=15} & \textbf{k=30} \\
\midrule
cube-single & $27.50 \pm 7.8$ & $29.25 \pm 3.8$ & $\mathbf{36.00 \pm 3.1}$ & $31.00 \pm 1.0$ & $30.75 \pm 5.2$ \\
cube-double & $11.50 \pm 7.0$ & $15.75 \pm 3.0$ & $\mathbf{21.00 \pm 5.2}$ & $\mathbf{21.00 \pm 5.1}$ & $9.75 \pm 0.4$ \\
cube-triple & $1.25 \pm 0.8$ & $3.25 \pm 2.2$ & $5.75 \pm 2.4$ & $8.75 \pm 1.4$ & $\mathbf{10.50 \pm 4.0}$ \\
cube-quadruple & $0.00 \pm 0.0$ & $0.00 \pm 0.0$ & $\mathbf{0.25 \pm 0.4}$ & $0.00 \pm 0.0$ & $\mathbf{0.25 \pm 0.4}$ \\
scene & $\mathbf{74.75 \pm 6.5}$ & $66.75 \pm 1.3$ & $70.75 \pm 1.7$ & $71.00 \pm 7.5$ & $50.25 \pm 3.6$ \\
puzzle-3x3 & $8.50 \pm 2.2$ & $11.75 \pm 3.0$ & $31.75 \pm 5.5$ & $\mathbf{50.50 \pm 10.7}$ & $40.00 \pm 12.0$ \\
puzzle-4x4 & $31.00 \pm 1.2$ & $32.00 \pm 3.6$ & $34.00 \pm 1.2$ & $\mathbf{38.00 \pm 2.6}$ & $29.25 \pm 2.8$ \\
puzzle-4x5 & $6.00 \pm 1.7$ & $\mathbf{6.75 \pm 1.6}$ & $3.75 \pm 0.4$ & $4.00 \pm 2.0$ & $2.75 \pm 0.8$ \\
puzzle-4x6 & $\mathbf{4.75 \pm 2.5}$ & $4.25 \pm 1.3$ & $3.25 \pm 1.3$ & $3.25 \pm 1.6$ & $1.75 \pm 1.3$ \\
\bottomrule
\end{tabular}
\vspace{3mm}
\caption{Sensitivity to horizon length $k$ on manipulation environments. Success rates (\%) averaged over 8 seeds with 95\% confidence intervals reported. Results within 95\% of the best are \textbf{bolded}.}
\label{tab:k_sweep_manipulation}
\end{table}
}

\newcommand{\visualresults}{ 
\begin{tabular}{lcc}
\toprule
Task & HIQL+CARL & HIQL \\
\midrule \texttt{visual-antmaze-medium} & $\mathbf{97.3 \pm 2.7}$ & $\mathbf{93 \pm 6.4}$ \\
\texttt{visual-antmaze-large} & $\mathbf{85.5 \pm 5.6}$ & $53 \pm 14.3$ \\
\texttt{visual-antmaze-giant} & $\mathbf{43.0 \pm 6.2}$ & $6 \pm 6.4$ \\
\texttt{visual-antmaze-teleport} & $\mathbf{45.3 \pm 2.7}$ & $37 \pm 3.2$ \\
\midrule
\texttt{visual-humanoidmaze-medium} & $\mathbf{2.3 \pm 2.4}$ & $0 \pm 0.0$ \\
\texttt{visual-humanoidmaze-large} & $\mathbf{0.0 \pm 0.0}$ & $\mathbf{0 \pm 0.0}$ \\
\texttt{visual-humanoidmaze-giant} & $\mathbf{0.0 \pm 0.0}$ & $\mathbf{0 \pm 0.0}$ \\
\midrule
\texttt{visual-cube-single} & $87.8 \pm 8.1$ & $\mathbf{89 \pm 0.0}$ \\
\texttt{visual-cube-double} & $\mathbf{41.8 \pm 9.4}$ & $39 \pm 3.2$ \\
\texttt{visual-cube-triple} & $\mathbf{24.8 \pm 2.7}$ & $21 \pm 0.0$ \\
\texttt{visual-cube-quadruple} & $\mathbf{14.8 \pm 4.5}$ & $14 \pm 1.6$ \\
\midrule
\texttt{visual-puzzle-3x3} & $\mathbf{75.5 \pm 4.6}$ & $73 \pm 12.7$ \\
\texttt{visual-puzzle-4x4} & $\mathbf{83.5 \pm 7.0}$ & $60 \pm 65.2$ \\
\texttt{visual-puzzle-4x5} & $\mathbf{18.0 \pm 3.7}$ & $13 \pm 14.3$ \\
\texttt{visual-puzzle-4x6} & $\mathbf{15.8 \pm 8.0}$ & $9 \pm 9.5$ \\
\midrule
\texttt{visual-scene} & $\mathbf{54.3 \pm 2.1}$ & $49 \pm 6.4$ \\
\midrule
\texttt{win-rate} & $\mathbf{13 / 14}$ & $1 / 14$ \\
\bottomrule \end{tabular}
}

\newcommand{\mergedresults}{
\begin{tabular}{l cc | cc | cccc}
\toprule
& \multicolumn{4}{c|}{\textbf{(a) HRL Comparisons}\phantomsubcaption\label{tab:hrl_comparison}}
& \multicolumn{4}{c}{\textbf{(b) \methodname{} Ablations}\phantomsubcaption\label{tab:carl_ablations}} \\
\cmidrule(lr){1-5} \cmidrule(lr){6-9}
\makecell[t]{Task} & \makecell[t]{HIQL+\methodname} & \makecell[t]{HIQL} & \makecell[t]{HGCBC+\methodname} & \makecell[t]{HGCBC} & \makecell[t]{\methodname} & \makecell[t]{Single-Action\\\methodname} & \makecell[t]{Multi-Action\\IDM} & \makecell[t]{Single-Action\\IDM} \\
\midrule
\texttt{pointmaze-medium} & \textbf{84.4 $\pm$ 3.1} & 79 $\pm$ 4.3 & \textbf{13.6 $\pm$ 1.8} & 0.0 $\pm$ 0.0 & \textbf{84.4 $\pm$ 3.1} & \textbf{86.6 $\pm$ 3.8} & 67.6 $\pm$ 4.0 & 74.1 $\pm$ 5.4 \\
\texttt{pointmaze-large} & \textbf{75.4 $\pm$ 3.1} & 58 $\pm$ 4.3 & \textbf{19.5 $\pm$ 3.8} & 0.4 $\pm$ 0.3 & \textbf{75.4 $\pm$ 3.1} & \textbf{74.7 $\pm$ 8.0} & 58.1 $\pm$ 9.5 & 62.6 $\pm$ 7.1 \\
\texttt{pointmaze-giant} & \textbf{64.5 $\pm$ 15.8} & 46 $\pm$ 7.6 & \textbf{0.0 $\pm$ 0.0} & \textbf{0.0 $\pm$ 0.0} & \textbf{64.5 $\pm$ 15.8} & 53.6 $\pm$ 13.2 & 22.7 $\pm$ 7.3 & 18.2 $\pm$ 9.2 \\
\texttt{pointmaze-teleport} & \textbf{31.4 $\pm$ 14.0} & 18 $\pm$ 3.3 & \textbf{29.7 $\pm$ 1.8} & 16.7 $\pm$ 1.2 & \textbf{31.4 $\pm$ 14.0} & 25.0 $\pm$ 8.0 & 18.0 $\pm$ 4.7 & 19.4 $\pm$ 6.4 \\
\midrule
\texttt{antmaze-medium} & \textbf{97.9 $\pm$ 0.9} & \textbf{96 $\pm$ 0.9} & \textbf{73.1 $\pm$ 1.5} & 59.8 $\pm$ 1.6 & \textbf{97.9 $\pm$ 0.9} & \textbf{95.5 $\pm$ 2.4} & \textbf{96.9 $\pm$ 1.7} & \textbf{96.1 $\pm$ 1.2} \\
\texttt{antmaze-large} & \textbf{91.9 $\pm$ 2.4} & \textbf{91 $\pm$ 1.7} & \textbf{64.9 $\pm$ 0.9} & 59.3 $\pm$ 1.9 & \textbf{91.9 $\pm$ 2.4} & 86.4 $\pm$ 6.1 & \textbf{91.5 $\pm$ 3.5} & \textbf{91.5 $\pm$ 1.2} \\
\texttt{antmaze-giant} & \textbf{75.2 $\pm$ 3.8} & 65 $\pm$ 4.3 & \textbf{18.4 $\pm$ 1.3} & 8.5 $\pm$ 0.8 & \textbf{75.2 $\pm$ 3.8} & 47.1 $\pm$ 7.3 & 70.0 $\pm$ 4.3 & \textbf{72.3 $\pm$ 3.5} \\
\texttt{antmaze-teleport} & \textbf{41.4 $\pm$ 2.1} & \textbf{42 $\pm$ 2.6} & \textbf{38.5 $\pm$ 1.3} & \textbf{36.9 $\pm$ 1.9} & 41.4 $\pm$ 2.1 & 42.5 $\pm$ 4.5 & \textbf{46.6 $\pm$ 2.4} & \textbf{47.0 $\pm$ 4.5} \\
\midrule
\texttt{humanoidmaze-medium} & \textbf{90.5 $\pm$ 2.4} & \textbf{89 $\pm$ 1.7} & \textbf{44.4 $\pm$ 1.3} & 32.4 $\pm$ 0.9 & \textbf{90.5 $\pm$ 2.4} & \textbf{88.9 $\pm$ 4.5} & \textbf{86.5 $\pm$ 3.1} & \textbf{87.1 $\pm$ 2.8} \\
\texttt{humanoidmaze-large} & \textbf{58.3 $\pm$ 3.8} & 49 $\pm$ 3.3 & \textbf{31.0 $\pm$ 0.8} & 22.6 $\pm$ 0.8 & \textbf{58.3 $\pm$ 3.8} & 51.1 $\pm$ 3.3 & 45.9 $\pm$ 3.3 & 50.0 $\pm$ 4.0 \\
\texttt{humanoidmaze-giant} & \textbf{27.2 $\pm$ 3.1} & 12 $\pm$ 3.3 & \textbf{20.0 $\pm$ 1.1} & 9.5 $\pm$ 1.3 & \textbf{27.2 $\pm$ 3.1} & 18.1 $\pm$ 3.8 & 17.0 $\pm$ 5.2 & 16.3 $\pm$ 3.1 \\
\midrule
\texttt{antsoccer-medium} & \textbf{12.8 $\pm$ 2.4} & \textbf{13 $\pm$ 1.7} & \textbf{7.0 $\pm$ 0.4} & 5.9 $\pm$ 0.7 & \textbf{12.8 $\pm$ 2.4} & 5.6 $\pm$ 1.7 & 10.9 $\pm$ 2.8 & 12.1 $\pm$ 2.6 \\
\texttt{antsoccer-arena} & \textbf{63.7 $\pm$ 3.1} & 58 $\pm$ 1.7 & \textbf{24.9 $\pm$ 1.6} & 12.5 $\pm$ 0.8 & \textbf{63.7 $\pm$ 3.1} & 56.0 $\pm$ 7.3 & 56.9 $\pm$ 4.0 & \textbf{63.0 $\pm$ 5.2} \\
\midrule
\texttt{cube-single} & \textbf{32.8 $\pm$ 3.1} & 15 $\pm$ 2.6 & \textbf{6.9 $\pm$ 0.7} & 4.4 $\pm$ 0.8 & \textbf{32.8 $\pm$ 3.1} & 30.5 $\pm$ 6.9 & 25.9 $\pm$ 4.3 & 22.4 $\pm$ 3.5 \\
\texttt{cube-double} & \textbf{23.4 $\pm$ 4.5} & 6 $\pm$ 1.7 & \textbf{1.3 $\pm$ 0.3} & 1.1 $\pm$ 0.3 & \textbf{23.4 $\pm$ 4.5} & 18.9 $\pm$ 4.0 & 5.4 $\pm$ 2.6 & 5.4 $\pm$ 1.2 \\
\texttt{cube-triple} & \textbf{15.2 $\pm$ 3.5} & 3 $\pm$ 0.9 & \textbf{2.1 $\pm$ 0.6} & 0.4 $\pm$ 0.3 & 15.2 $\pm$ 3.5 & \textbf{24.4 $\pm$ 4.0} & 6.4 $\pm$ 2.1 & 9.0 $\pm$ 4.0 \\
\texttt{cube-quadruple} & \textbf{0.1 $\pm$ 0.2} & 0 $\pm$ 0.0 & \textbf{0.1 $\pm$ 0.1} & 0.0 $\pm$ 0.0 & 0.1 $\pm$ 0.2 & \textbf{1.8 $\pm$ 1.7} & 0.4 $\pm$ 0.7 & 0.3 $\pm$ 0.5 \\
\midrule
\texttt{puzzle-3x3} & \textbf{45.5 $\pm$ 7.3} & 12 $\pm$ 1.7 & \textbf{6.0 $\pm$ 0.9} & 4.1 $\pm$ 0.3 & \textbf{45.5 $\pm$ 7.3} & 15.0 $\pm$ 4.3 & 27.1 $\pm$ 5.2 & 28.1 $\pm$ 3.8 \\
\texttt{puzzle-4x4} & \textbf{35.2 $\pm$ 7.1} & 7 $\pm$ 1.7 & \textbf{5.9 $\pm$ 0.9} & 0.3 $\pm$ 0.2 & \textbf{35.2 $\pm$ 7.1} & 22.5 $\pm$ 4.0 & 14.2 $\pm$ 3.1 & 13.7 $\pm$ 2.4 \\
\texttt{puzzle-4x5} & \textbf{4.2 $\pm$ 2.4} & \textbf{4 $\pm$ 0.9} & \textbf{1.6 $\pm$ 0.6} & 0.4 $\pm$ 0.2 & 4.2 $\pm$ 2.4 & \textbf{5.0 $\pm$ 1.4} & 3.0 $\pm$ 1.9 & 3.1 $\pm$ 0.9 \\
\texttt{puzzle-4x6} & \textbf{3.6 $\pm$ 2.1} & 3 $\pm$ 0.9 & \textbf{0.6 $\pm$ 0.3} & 0.3 $\pm$ 0.2 & 3.6 $\pm$ 2.1 & \textbf{3.9 $\pm$ 1.7} & 2.4 $\pm$ 1.4 & 2.3 $\pm$ 1.7 \\
\midrule
\texttt{scene} & \textbf{70.5 $\pm$ 5.2} & 38 $\pm$ 2.6 & \textbf{15.8 $\pm$ 1.0} & 8.5 $\pm$ 1.1 & \textbf{70.5 $\pm$ 5.2} & \textbf{69.8 $\pm$ 6.4} & 55.7 $\pm$ 5.7 & 64.3 $\pm$ 3.1 \\
\midrule
\texttt{\textbf{win-rate}} & \textbf{20/22} &	2/22 & \textbf{21/21} & 0/21 & \textbf{16/22} & 5/22 & 0/22 & 1/22 \\
\bottomrule
\end{tabular}
}

\newcommand{\repdimresults}{
\begin{tabular}{l ccc | ccc}
\toprule
& \multicolumn{3}{c|}{\textbf{HIQL Variants}\phantomsubcaption\label{tab:hiql_repdim}}
& \multicolumn{3}{c}{\textbf{HGCBC Variants}\phantomsubcaption\label{tab:hgcbc_repdim}} \\
\cmidrule(lr){1-4} \cmidrule(lr){5-7}
\makecell[t]{Task} & \makecell[t]{HIQL+\methodname} & \makecell[t]{HIQL\\(rep\_dim=100)} & \makecell[t]{HIQL} & \makecell[t]{HGCBC+\methodname} & \makecell[t]{HGCBC\\(rep\_dim=100)} & \makecell[t]{HGCBC} \\
\midrule
\texttt{pointmaze-medium} & \textbf{84.4 $\pm$ 3.1} & 65.5 $\pm$ 7.3 & 79 $\pm$ 4.3 & \textbf{13.6 $\pm$ 1.8} & 0.0 $\pm$ 0.0 & 0.0 $\pm$ 0.0 \\
\texttt{pointmaze-large} & \textbf{75.4 $\pm$ 3.1} & 60.1 $\pm$ 10.2 & 58 $\pm$ 4.3 & \textbf{19.5 $\pm$ 3.8} & 0.3 $\pm$ 0.2 & 0.4 $\pm$ 0.3 \\
\texttt{pointmaze-giant} & \textbf{64.5 $\pm$ 15.8} & 21.7 $\pm$ 6.9 & 46 $\pm$ 7.6 & \textbf{0.0 $\pm$ 0.0} & \textbf{0.0 $\pm$ 0.0} & \textbf{0.0 $\pm$ 0.0} \\
\texttt{pointmaze-teleport} & \textbf{31.4 $\pm$ 14.0} & 16.0 $\pm$ 4.5 & 18 $\pm$ 3.3 & \textbf{29.7 $\pm$ 1.8} & 12.6 $\pm$ 1.1 & 16.7 $\pm$ 1.2 \\
\midrule
\texttt{antmaze-medium} & \textbf{97.9 $\pm$ 0.9} & \textbf{96.5 $\pm$ 1.7} & \textbf{96 $\pm$ 0.9} & \textbf{73.1 $\pm$ 1.5} & 60.7 $\pm$ 2.2 & 59.8 $\pm$ 1.6 \\
\texttt{antmaze-large} & \textbf{91.9 $\pm$ 2.4} & \textbf{89.9 $\pm$ 3.3} & \textbf{91 $\pm$ 1.7} & \textbf{64.9 $\pm$ 0.9} & 48.0 $\pm$ 0.9 & 59.3 $\pm$ 1.9 \\
\texttt{antmaze-giant} & \textbf{75.2 $\pm$ 3.8} & 68.8 $\pm$ 5.4 & 65 $\pm$ 4.3 & \textbf{18.4 $\pm$ 1.3} & 4.4 $\pm$ 1.1 & 8.5 $\pm$ 0.8 \\
\texttt{antmaze-teleport} & 41.4 $\pm$ 2.1 & \textbf{45.6 $\pm$ 6.6} & 42 $\pm$ 2.6 & \textbf{38.5 $\pm$ 1.3} & 34.6 $\pm$ 2.1 & \textbf{36.9 $\pm$ 1.9} \\
\midrule
\texttt{humanoidmaze-medium} & \textbf{90.5 $\pm$ 2.4} & \textbf{87.6 $\pm$ 4.5} & \textbf{89 $\pm$ 1.7} & \textbf{44.4 $\pm$ 1.3} & 40.8 $\pm$ 1.8 & 32.4 $\pm$ 0.9 \\
\texttt{humanoidmaze-large} & \textbf{58.3 $\pm$ 3.8} & 52.9 $\pm$ 5.2 & 49 $\pm$ 3.3 & \textbf{31.0 $\pm$ 0.8} & 19.8 $\pm$ 1.7 & 22.6 $\pm$ 0.8 \\
\texttt{humanoidmaze-giant} & \textbf{27.2 $\pm$ 3.1} & 22.8 $\pm$ 4.5 & 12 $\pm$ 3.3 & \textbf{20.0 $\pm$ 1.1} & 12.0 $\pm$ 1.3 & 9.5 $\pm$ 1.3 \\
\midrule
\texttt{antsoccer-medium} & \textbf{12.8 $\pm$ 2.4} & 11.7 $\pm$ 2.8 & \textbf{13 $\pm$ 1.7} & \textbf{7.0 $\pm$ 0.4} & 4.9 $\pm$ 1.0 & 5.9 $\pm$ 0.7 \\
\texttt{antsoccer-arena} & \textbf{63.7 $\pm$ 3.1} & 57.9 $\pm$ 3.1 & 58 $\pm$ 1.7 & \textbf{24.9 $\pm$ 1.6} & 15.2 $\pm$ 1.2 & 12.5 $\pm$ 0.8 \\
\midrule
\texttt{cube-single} & \textbf{32.8 $\pm$ 3.1} & 19.9 $\pm$ 4.0 & 15 $\pm$ 2.6 & \textbf{6.9 $\pm$ 0.7} & 5.6 $\pm$ 1.1 & 4.4 $\pm$ 0.8 \\
\texttt{cube-double} & \textbf{23.4 $\pm$ 4.5} & 4.5 $\pm$ 2.4 & 6 $\pm$ 1.7 & \textbf{1.3 $\pm$ 0.3} & \textbf{1.3 $\pm$ 0.5} & 1.1 $\pm$ 0.3 \\
\texttt{cube-triple} & \textbf{15.2 $\pm$ 3.5} & 3.3 $\pm$ 1.2 & 3 $\pm$ 0.9 & \textbf{2.1 $\pm$ 0.6} & 0.0 $\pm$ 0.0 & 0.4 $\pm$ 0.3 \\
\texttt{cube-quadruple} & 0.1 $\pm$ 0.2 & \textbf{0.3 $\pm$ 0.5} & 0 $\pm$ 0.0 & \textbf{0.1 $\pm$ 0.1} & 0.0 $\pm$ 0.0 & 0.0 $\pm$ 0.0 \\
\midrule
\texttt{puzzle-3x3} & 45.5 $\pm$ 7.3 & \textbf{63.4 $\pm$ 8.3} & 12 $\pm$ 1.7 & \textbf{6.0 $\pm$ 0.9} & 4.0 $\pm$ 0.8 & 4.1 $\pm$ 0.3 \\
\texttt{puzzle-4x4} & \textbf{35.2 $\pm$ 7.1} & 4.3 $\pm$ 2.6 & 7 $\pm$ 1.7 & \textbf{5.9 $\pm$ 0.9} & 0.3 $\pm$ 0.2 & 0.3 $\pm$ 0.2 \\
\texttt{puzzle-4x5} & 4.2 $\pm$ 2.4 & \textbf{5.2 $\pm$ 1.9} & 4 $\pm$ 0.9 & \textbf{1.6 $\pm$ 0.6} & 1.1 $\pm$ 0.5 & 0.4 $\pm$ 0.2 \\
\texttt{puzzle-4x6} & \textbf{3.6 $\pm$ 2.1} & 2.5 $\pm$ 1.7 & 3 $\pm$ 0.9 & \textbf{0.6 $\pm$ 0.3} & \textbf{0.6 $\pm$ 0.2} & 0.3 $\pm$ 0.2 \\
\midrule
\texttt{scene} & \textbf{70.5 $\pm$ 5.2} & 41.0 $\pm$ 2.6 & 38 $\pm$ 2.6 & \textbf{15.8 $\pm$ 1.0} & 7.7 $\pm$ 1.3 & 8.5 $\pm$ 1.1 \\
\midrule
\texttt{\textbf{win-rate}} & \textbf{17/22} & 4/22 & 1/22 & \textbf{19/19} & 0/19 & 0/19 \\
\bottomrule
\end{tabular}
}

\newcommand{\methodname}{CARL\xspace}
\newcommand{\singleactionmethodname}{Single-Action \methodname\xspace}

\newcommand{\highlight}[1]{%
  \ifmmode
    \begingroup\setlength{\fboxsep}{0pt}\colorbox{cyan!15}{$#1$}\endgroup%
  \else
    \begingroup\setlength{\fboxsep}{0pt}\colorbox{cyan!15}{#1}\endgroup%
  \fi
}

\usepackage{etoolbox}
\hypersetup{
  colorlinks=true,
  citecolor=NavyBlue,
  linkcolor=NavyBlue,
  urlcolor=green
}

\usepackage{silence}
\usepackage{amsmath}
\usepackage{amssymb}
\usepackage{mathtools}
\usepackage{amsthm}

\usepackage[capitalize,noabbrev]{cleveref}

\theoremstyle{plain}
\newtheorem{theorem}{Theorem}[section]

\theoremstyle{definition}
\newtheorem{definition}[theorem]{Definition}

\theoremstyle{remark}

 \usepackage[preprint]{neurips_2026}

\usepackage[utf8]{inputenc} 
\usepackage[T1]{fontenc}    
\usepackage{hyperref}       
\usepackage{url}            
\usepackage{booktabs}       
\usepackage{amsfonts}       
\usepackage{nicefrac}       
\usepackage{microtype}      
\usepackage{xcolor}         

\title{Exploiting Local Dynamics Regularity for Reusable Skills in Offline Hierarchical RL}

\author{
  Sarthak Dayal\thanks{Equal contribution.} \\
  Department of Computer Science \\
  University of Texas at Austin \\
  \texttt{sarthak@utexas.edu}
  \And
  Abhinav Peri\hyperlink{Hfootnote.1}{\textsuperscript{*}} \\
  Department of Computer Science \\
  University of Texas at Austin \\
  \texttt{app2452@utexas.edu}
  \AND
  Carl Qi \\
  UT Austin \\
  \And
  Claas Voelcker \\
  UT Austin \\
  \And
  Alexander Levine \\
  OpenAI \\
  \And
  Caleb Chuck \\
  UT Austin \\
  \And
  Amy Zhang \\
  UT Austin \\
}

\begin{document}

\maketitle

\begin{abstract}

Hierarchical Reinforcement Learning (HRL) promises to solve long-horizon Reinforcement Learning (RL) tasks more efficiently than non-hierarchical counterparts by discovering and reusing temporally-extended skills. 
However, obtaining skills that are actually reusable remains an open challenge.
Towards this end, we focus on abstractions that exploit the intuition of local dynamics: local transitions in different global contexts require similar kinds of action sequences. By aligning these contexts with the action sequences they require, we are able to learn which skills to reuse and where to reuse them. In principle, this information should benefit many HRL algorithms, where high-level policies have to reason about the low-level skills they use. The resulting algorithm CARL (\textbf{C}ontrastive \textbf{A}ction-based \textbf{R}epresentations for Reusable \textbf{L}ocal Control) shows both qualitative clustering of meaningful skills in complex humanoid environments and improved downstream performance on the OGBench benchmark when integrated with HIQL. We visualize additional results and video rollouts on our accompanying website.\footnote{\url{https://sites.google.com/view/behavior-rep/home}}

\end{abstract}

\section{Introduction}
\label{sec:intro}

Hierarchical Reinforcement Learning (HRL) methods offer a framework~\citep{sutton1999options} for abstracting low-level policies or action sequences---commonly referred to as \emph{skills}---so that they can be reused to accomplish complex tasks. However, this promise of reusable skills has often been outweighed by the limitations of existing approaches~\citep{zhang2021hrllimitations}, which include training instability when co-learning skills with the policies that use those skills~\citep{levy2019hac, wang2025hierarchical}, and failure to leverage reusability. One promising direction involves representing low-level skills as goal-conditioned policies~\citep{park2024hiql, hafner2022deephierarchicalplanningpixels, eysenbach2019search}, which decouples low- and high-level policy training. However, representing skills via a goal-conditioned low-level policy loses many of the desirable properties of skills, such as temporal abstraction and consistent reusability. In this work, we formalize the idea of local dynamics---the idea that many local regions of the state space share similar dynamics---to enable reuse of low-level goal-conditioned skills across the state space.

\begin{figure}[t]
\centering
\includegraphics[width=1.0\linewidth]{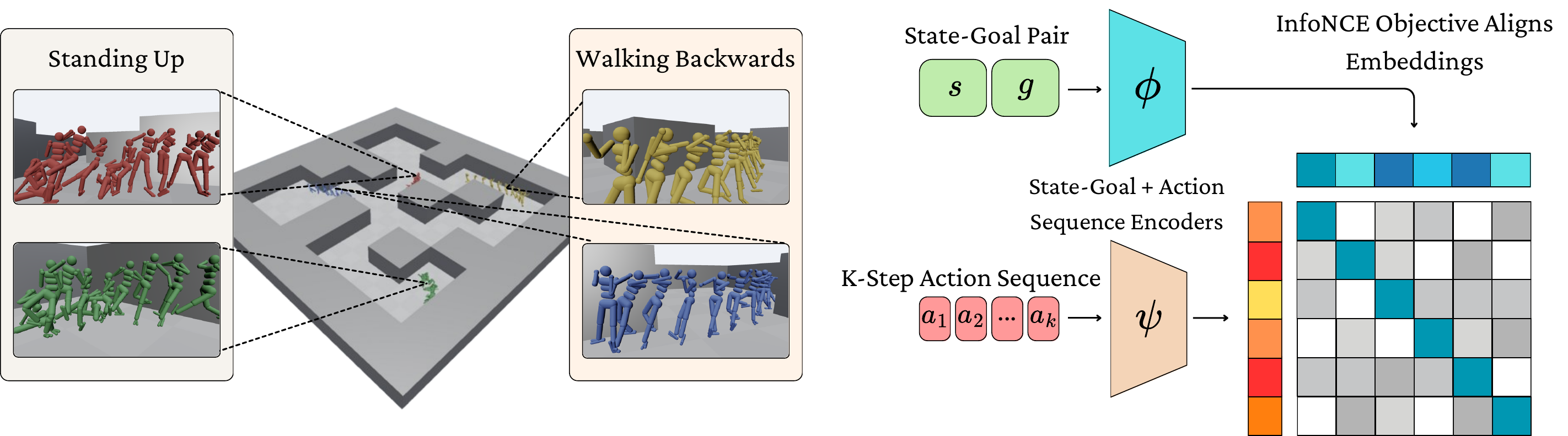}
\caption{Learning state--goal representations for skill reuse. Humanoid poses left-to-right in each pane reflect progress in time. \methodname learns a representation $\phi(s,g)$ that clusters state--goal pairs if they admit the same kinds of $k$-step action sequences. \methodname clusters ``walking backward" and ``standing up" behaviors for humanoids situated in globally different parts of the maze.}
\label{fig:method}
\end{figure}

To formalize the idea of local dynamics we define local, finite-horizon MDPs for each state to capture its surrounding states and their dynamics. We then draw on bisimulation as a principled way to define state equivalences based on these MDPs. To derive a practical algorithm from this principle, we leverage the idea of \emph{behavioral similarity}: two state-goal pairs $(s_1, g_1)$ and $(s_2, g_2)$ are considered similar when a policy can use the same skill $(a_1, a_2, a_3, \dots, a_k)$ to achieve the transition from $s_1$ to $g_1$ and from $s_2$ to $g_2$. Our hypothesis is that behavioral similarity will naturally emerge when many states share local dynamics structure, making it a useful heuristic for understanding when skills can be reused. 
Our method is built upon these insights and leverages a contrastive learning objective to align state-goal pairs with the action sequences that achieve them. 

We introduce \methodname (\textbf{C}ontrastive \textbf{A}ction-based \textbf{R}epresentations for Reusable \textbf{L}ocal Control) to learn representations of local dynamics from offline datasets at a fixed skill horizon. By recognizing the underlying local dynamics structure, \methodname identifies where skills can be reused instead of forcing HRL methods to relearn low-level policies from scratch. We demonstrate this by integrating these abstractions into existing HRL algorithms, HIQL and HGCBC ~\citep{park2024hiql} 
which yields clear performance benefits on the OGBench benchmark.
Furthermore, our objective shapes the latent geometry so that state–goal pairs requiring similar local skills are embedded nearby, even when they occur in different regions of the environment. Figure~\ref{fig:method} illustrates this in a humanoid maze environment, where \methodname groups skills together regardless of where they occur in the maze. 


As noted, \methodname relies on an offline dataset and fixed horizon to capture local dynamics, which may constrain expressible skills when data coverage is poor or the horizon fails to capture useful structure. We include ablations analyzing the effects of coverage, imbalance, and horizon length. Overall, our work shows that local dynamics structure provides a principled basis for skill extraction and reuse, helping HRL reconnect with the temporal abstractions central to long-horizon decision-making.
\section{Related Work}
\subsection{Hierarchical Reinforcement Learning}
Hierarchical Reinforcement Learning studies how to solve long-horizon Reinforcement Learning (RL) tasks by introducing \emph{temporal abstraction}, which refers to high-level decisions that invoke temporally extended behaviors. One of the first formulations of this idea is the options framework, which models a skill as a policy with an initiation set and termination condition, enabling re-use of temporally extended actions within an SMDP view of control \citep{sutton1999options}. Early HRL work explored a variety of hierarchical decompositions and skill abstractions, including
feudal-style manager--worker architectures~\citep{dayan1992feudal}, value-function decomposition methods such as MAXQ~\citep{dietterich1999maxq}, and policy-constraining formalisms such as hierarchies of abstract machines (HAMs)~\citep{parr1997reinforcement}.

A core challenge in learning hierarchies \emph{end-to-end} is non-stationarity. As the low-level policy changes, the effective dynamics faced by the high-level policy shift, often destabilizing joint optimization. Many modern approaches therefore decouple the training of the hierarchy, by using subgoals (goals in state space) to train a local goal-conditioned policy while learning a high-level policy that proposes useful targets \citep{nachum2018dataefficienthierarchicalreinforcementlearning, levy2019hac}. Recent offline and model-based hierarchical methods further improve this paradigm by extracting effective policies from offline data or learned dynamics models \citep{hafner2022deephierarchicalplanningpixels, park2024hiql}. Although this paradigm substantially improves the training stability of HRL methods, reasoning directly in the global state–goal space makes it difficult for the low-level policy to recognize when the same behavior can be reused. To overcome this limitation, we focus on learning representations of state–goal pairs that highlight the short-horizon action structure and thus enable appropriate reuse of behaviors.

\subsection{Abstractions via Behavioral Equivalences and Invariances}

 In this section, we discuss methods that seek to generalize by learning behavior-preserving abstractions, where distinct states (or state-goal contexts) are mapped to similar representations when they are interchangeable \citep{agarwal2021behavioralsimilarity, hansen2022gcbisim, islam2023agentcontrollerrepresentationsprincipledoffline, ajay2021opal, park2025dualgoalrepresentations}. A common theme is to define an equivalence relation that preserves global decision-relevant structure: for example via value-, policy-, or model-based notions of similarity---and then learn representations that collapse equivalent situations while discarding irrelevant variation. This perspective is closely related to bisimulation-based ideas, which characterize when two states can be treated as equivalent without materially changing long-horizon outcomes \citep{zhang2021bisim,rudolph2024learningactionbasedrepresentationsusing,hansen2022gcbisim,castro2020scalable}.

Our work fits into this line of research, but fundamentally targets a complementary notion of equivalence grounded in \emph{local reuse}. Rather than requiring states to match in long-horizon value or reward structure, we ask when different state-goal pairs admit similar local goal-reaching behaviors, which allows us to cluster exactly the state-goal pairs which are the same from the perspective of short-horizon control, regardless of the long-horizon effects.

\subsection{Contrastive Representations for Control}

Contrastive Representation Learning provides a general mechanism for extracting structure in embeddings that score positive pairs higher than selected negative pairs, typically trained via objectives such as InfoNCE or NCE \citep{oord2019cpc, radford2021clip}. In RL, contrastive methods have been applied to learn task relevant structure and enable downstream generalization \citep{eysenbach2023crl, agarwal2021behavioralsimilarity, srinivas2020curl}. Particularly, Contrastive RL \citep{eysenbach2023crl} connects contrastive objectives to goal-conditioned RL by learning representations in which reaching a goal corresponds to matching future states under a learned similarity metric. Our method is inspired by this viewpoint, but captures a different structure: instead of learning representations that reflect goal reachability, we capture when similar action sequences are admitted by the state-goal transitions that the low-level policy aims to achieve. This greatly changes the contrastive structure compared to previous approaches that instead reason about the long-horizon effects of actions.

\section{Preliminaries}

\subsection{Offline Goal-Conditioned RL}
In this work, we investigate behavioral equivalences in the context of offline goal-conditioned RL (OGCRL), which is defined by the Markov decision process $\mathcal M \coloneqq (\mathcal S, \mathcal A, p, r, \gamma)$, where $\mathcal S$ is the state space, $\mathcal A$ is the action space, $p: \mathcal S \times \mathcal A \rightarrow \Delta(\mathcal S)$ is the transition function, and $\Delta(\mathcal S)$ is a distribution over states, $r: \mathcal S \times \mathcal G \rightarrow \mathbb R$ is a reward function, and $\gamma \in (0,1)$ is a discount factor. We consider the case where $\mathcal G \subseteq \mathcal S$. The objective is to learn a goal conditioned policy $\pi: \mathcal S \times \mathcal G \rightarrow \Delta(A)$ such that for time horizon $H$, the expected reward is maximized: $\max_\pi J(\pi) = \max_\pi \mathbb E_{g\sim \rho(g),\tau \sim \rho^\pi(\tau)}[\sum_{t=0}^H \gamma^t r(s_t,g)]$. We define trajectory $\tau = (s_0,a_0,  s_1,a_1,\hdots,s_H)$ to be a sequence of states with a fixed horizon $H$. Trajectories are sampled according to $\rho^\pi(\tau) = \mu(s_0)\Pi_{t=0}^{H-1} \pi(a\mid s_t,g) p(s_{t+1}|s_t,a_t)$, where $\mu$ is the initial distribution over states, and $\rho(g)$ is the goal distribution. We write $p(s' \mid s, \mathbf{a_k}) = \sum_{s_1,\ldots,s_{k-1} \in \mathcal{S}}
\prod_{i=0}^{k-1}
p(s_{i+1}\mid s_i,a_i),$ to mean the $k$-step transition function, where $\mathbf{a_k}$ denotes a $k$-step action sequence starting at state $s$, and $a_i$ denote individual actions in that sequence. We aim to use upper case letters for random variables, script upper case for spaces, and lower case for values and functions.

\subsection{Local Dynamics}
Here we formalize our intuition of local dynamics as a state equivalence relation. Specifically, we use the transition dynamics within a fixed horizon $k$ in order to characterize a state's local dynamics.

\begin{definition}[$k$-ball]
\label{def:k-ball}
For any $k \in \mathbb{N}$ and $s \in \mathcal{S}$, the \emph{$k$-ball} of $s$, denoted $\mathcal{B}_k(s)$, is the set of states reachable from $s$ within $k$ steps. Formally, $s' \in \mathcal{B}_k(s)$ if there exist $t \in \{0, 1, \dots, k\}$ and an action sequence $\mathbf{a_t}$ such that $p(s' \mid s, \mathbf{a_t}) > 0$.
\end{definition}

\begin{definition}[$k$-ball MDP]
\label{def:k-ball-mdp}
The \emph{$k$-ball MDP} rooted at $s$ is the finite-horizon MDP
\[
\mathcal{M}_s^{(k)} = \bigl(\mathcal{B}_k(s),\ \mathcal{A},\ p,\ r,\ k\bigr),
\]
whose transition dynamics $P$, action space $\mathcal{A}$, and reward $R$ are inherited from $\mathcal{M}$.
\end{definition}

We now use the $k$-ball MDP to formalize a notion of state similarity grounded in local dynamics. Prior work adapting bisimulation to RL \citep{ferns, zhang2021bisim, rudolph2024learningactionbasedrepresentationsusing} measure similarity between MDPs through the sequences of rewards or single-step controllability metrics under identical action sequences. Our formalism instead considers dynamics directly, capturing whether two MDPs are interchangeable for goal reaching behavior.

\begin{definition}[Dynamics Bisimilarity]
\label{def:dyn-bisim}

Let $\mathcal{M}_1 = (\mathcal{S}_1, \mathcal{A}, p_1, r_1)$ and $\mathcal{M}_2 = (\mathcal{S}_2, \mathcal{A}, p_2, r_2)$ be two MDPs sharing an action space $\mathcal{A}$. We say that $\mathcal{M}_1$ and $\mathcal{M}_2$ are \emph{dynamics-bisimilar} if there exists a total relation $B \subseteq \mathcal{S}_1 \times \mathcal{S}_2$ such that every $(x, x') \in B$ for all $a \in \mathcal{A}$ satisfies

\begin{align*}
\forall\, y \in \mathcal{S}_1,\ \exists\, y' \in \mathcal{S}_2 \text{ (and vice versa)} :\quad p_1(y \mid x, a) = p_2(y' \mid x', a), &\\
\forall\, C \in \mathcal{S}_1/B :\quad \bar{p_1}(C \mid x, a) = \bar{p_2}(B^{-1}(C) \mid x', a). &
\end{align*}

where $\mathcal{S}_1/B$ is the partition of $\mathcal{S}_1$ induced by $B$ (the set of equivalence classes $C$ of $B$-related states in $\mathcal{S}_1$), $B^{-1}(C) = \{y' \in \mathcal{S}_2 : \exists\, y \in C,\ (y, y') \in B\}$ is the corresponding subset of $\mathcal{S}_2$, and $\bar{p}(C \mid s, a) = \sum_{s' \in C} p(s' \mid s, a)$ extends $p$ to subsets.

\end{definition}

We say that two states $s$ and $s'$ share local dynamics if their corresponding $k$-ball MDPs, $\mathcal{M}_s^{(k)}$ and $\mathcal{M}_{s'}^{(k)}$, are \emph{dynamics-bisimilar}.

\section{Method}
\subsection{Learning Representation for Skill Reuse}
\label{sec:method-setup}

Although our state equivalence relates local dynamics, directly estimating this quantity in a practical algorithm for skill-reuse is challenging because it would require constructing the $k$-ball MDP around each state and comparing their transition structure under all relevant actions. In continuous, high-dimensional domains, the $k$-ball MDP is especially hard to obtain and enumerate for comparison. In the offline setting, this problem is compounded by the fact that we only observe a finite set of trajectories, which may sparsely cover the relevant local transitions. As a result, we seek an approximation that preserves the key ideas from our formalism while remaining simple to implement. 

Our core insight is that offline datasets with near-expert behavior can reveal local dynamics structure without requiring us to directly model dynamics-bisimulation relations as described in Definition~\ref{def:dyn-bisim}. Intuitively, when two states have dynamics-bisimilar $k$-ball MDPs, the same $k$-step action sequence accomplishes equivalent transitions from both. A near-expert data-collecting policy will exploit this, leaving repeated action-sequence structure as a footprint in the dataset. We call this footprint \emph{behavioral similarity}: two state-goal pairs $(s_1, g_1)$ and $(s_2, g_2)$ are behaviorally similar if they are solved by similar $k$-step action sequences in the dataset. This yields a much simpler heuristic that lets us identify skill-reuse opportunities directly from the dataset.


This motivates our method \methodname (\textbf{C}ontrastive \textbf{A}ction-based \textbf{R}epresentations for Reusable \textbf{L}ocal Control) which learns to relate skills  with the global contexts in which they are used. To do this, we sample batches of size $B$ from a diverse offline dataset to learn representations $\phi(s,g_k)$ and $\psi(\mathbf{a_k})$ contrastively such that the loss below is minimized. 

\begin{equation}
\label{eq:infonce}
\begin{aligned}
&\mathcal{L}_{\mathrm{InfoNCE}}\!\big(
\{(s^i, g_k^i, \mathbf{a_k}^i)\}_{i=1}^B;\, \phi, \psi
\big)
= -\frac{1}{B}\sum_{i=1}^B 
\log
\frac{
\exp\!\left(\langle \phi(s^i, g_k^i), \psi(\mathbf{a_k}^i)\rangle / \tau\right)
}{
\sum_{j=1}^B \exp\!\left(\langle \phi(s^i, g_k^i), \psi(\mathbf{a_k}^j)\rangle / \tau\right)
}.
\end{aligned}
\end{equation}

Algorithm~\ref{alg:lasr} summarizes the training procedure, which alternates between training the encoders via contrastive loss and training a goal-conditioned policy in the learned representation space.
\begin{figure}[t]
\begin{minipage}[t]{0.43\linewidth}
\begin{algorithm}[H]
\small
\caption{\methodname}
\LinesNumbered
\label{alg:lasr}
\textbf{Input:} Offline dataset $\mathcal{D}$, horizon $k$, temperature $\tau$\;
\textbf{Initialize:} Encoders $\phi(s,g)$ and $\psi(\mathbf{a_k})$\;
\textbf{Initialize:} Goal-conditioned learner $\pi(\cdot \mid \phi(s,g))$\;
\BlankLine
\While{not converged}{
    Sample batch $b \sim \mathcal{D}$ of $k$-step segments $(s_t,\mathbf{a_k},s_{t+k})$\;
    Train encoders: $\mathcal{L}_{\mathrm{InfoNCE}}(b; \phi,\psi)$\tcp*[r]{Eq.(\ref{eq:infonce})}
    Train policy: $\mathbb{E}_{b}[J(\pi\mid \phi)]$\;
}
\end{algorithm}
\end{minipage}%
\hspace{0.05\linewidth}
\begin{minipage}[t]{0.52\linewidth}
\begin{algorithm}[H]
\small
\caption{HIQL + \methodname\ (changes in blue)}
\label{alg:cotrain_compact}
\textbf{Input:} dataset $\mathcal{D}$, horizon $k$, temperature $\tau$\;
\textbf{Initialize:} value $V_{\theta_V}(s,\phi(s,g))$, policies $\pi^h_{\theta_h}(\cdot\mid s,g)$, $\pi^\ell_{\theta_\ell}(\cdot\mid s,\phi(s,s'))$\;
\highlight{\textbf{Initialize:} encoders $\phi_{\theta_\phi}(s,s')$, $\psi_{\theta_\psi}(a_{\tau})$}\;
\BlankLine
\While{not converged}{
    Sample minibatch $(s_t, a_{t:t+k-1}, s_{t+k}, g) \sim \mathcal D$, indices $j \sim \mathrm{Unif}[k]$\;
    Train encoders:\\
    \hspace*{1.5em}\highlight{$\mathcal{L}_{\mathrm{InfoNCE}}(\{s_t^i,s_{t+j}^i,a_{t:t+k-1}^i\}_{i}; \phi_{\theta_\phi},\psi_{\theta_\psi})$}\;
    Update value: $V_{\theta_V}(s_t, \phi_{\theta_\phi}(s_t, g))$\;
    \highlight{$z_t^\star \leftarrow \phi_{\theta_\phi}(s_t,s_{t+k})$}\tcp*[r]{\scriptsize embed subgoal}
    Train high-level: fit $\pi^h_{\theta_h}(\highlight{z_t^\star}\mid s_t,g)$ (AWR)\;
    Train low-level: fit $\pi^\ell_{\theta_\ell}(a_t\mid s_t,\highlight{z_t^\star})$ (AWR)\;
}
\end{algorithm}
\vspace{-5mm}
\end{minipage}
\end{figure}
\subsection{Integration with Hierarchical Offline RL}
\label{sec:method-hrl}


Aligning state-goal pairs with action sequences that achieve the transition between them induces a behaviorally structure representation: state--goal pairs that can be solved by similar $k$-step action sequences are embedded nearby, regardless of where they occur in the state space. This representation structure is well suited as an input to the low-level policy in subgoal-based HRL algorithms, where a policy can reuse the same skill in different contexts rather than relearning it separately in each region.

Following this intuition, we integrate \methodname into HIQL \citep{park2024hiql}, a recent and competitive hierarchical offline RL algorithm, by co-training its subgoal representation with \methodname's objective. This introduces \methodname's skill reuse benefits into the input space of the low-level policy and value function, and the output space of the high-level policy. These changes are highlighted in blue in Algorithm ~\ref{alg:cotrain_compact}. We provide further details on the specific sampling and training procedures that we used in Appendix~\ref{app:training}, specific hyperparameter changes we made in Appendix~\ref{app:impl}, and present experiments analyzing these design choices in Appendix~\ref{app:additional_exps}.

We also modify the HGCBC (\textbf{H}ierarchical \textbf{G}oal-\textbf{C}onditioned \textbf{B}ehavior \textbf{C}loning) algorithm used as a baseline in \cite{park2024hiql} to assess the generality of \methodname's benefits to HRL algorithms. Similarly to HIQL, we change the low-level policy inputs and high-level policy outputs to use \methodname's representation for the subgoal space. Section~\ref{sec:exp} provides strong empirical evidence for our generality claim: HIQL+\methodname achieves win rates of 20/22 on state-based and 13/14 on image-based tasks against HIQL, and HGCBC+\methodname achieves 21/21 on state-based tasks against HGCBC.

\section{Experiments}
\label{sec:exp}

Our experiments analyze both the embedding structure produced by \methodname and its effect on downstream performance when integrated into an existing HRL method. The following sections seek to address the questions below. 

\begin{enumerate}
    \item [Q1.] Does \methodname's representation structure enable skills to be reused?
    \item [Q2.] How does \methodname benefit HRL methods in downstream tasks?
    \item [Q3.] What are the important components for capturing behavioral similarity?
    \item [Q4.] How does \methodname organize different skills in its representation structure?
\end{enumerate}

\subsection{Skill Reuse in Toy Environments}
\label{sec:5rooms}

\begin{wrapfigure}{r}{0.45\textwidth}
    \centering
    \includegraphics[width=0.42\textwidth]{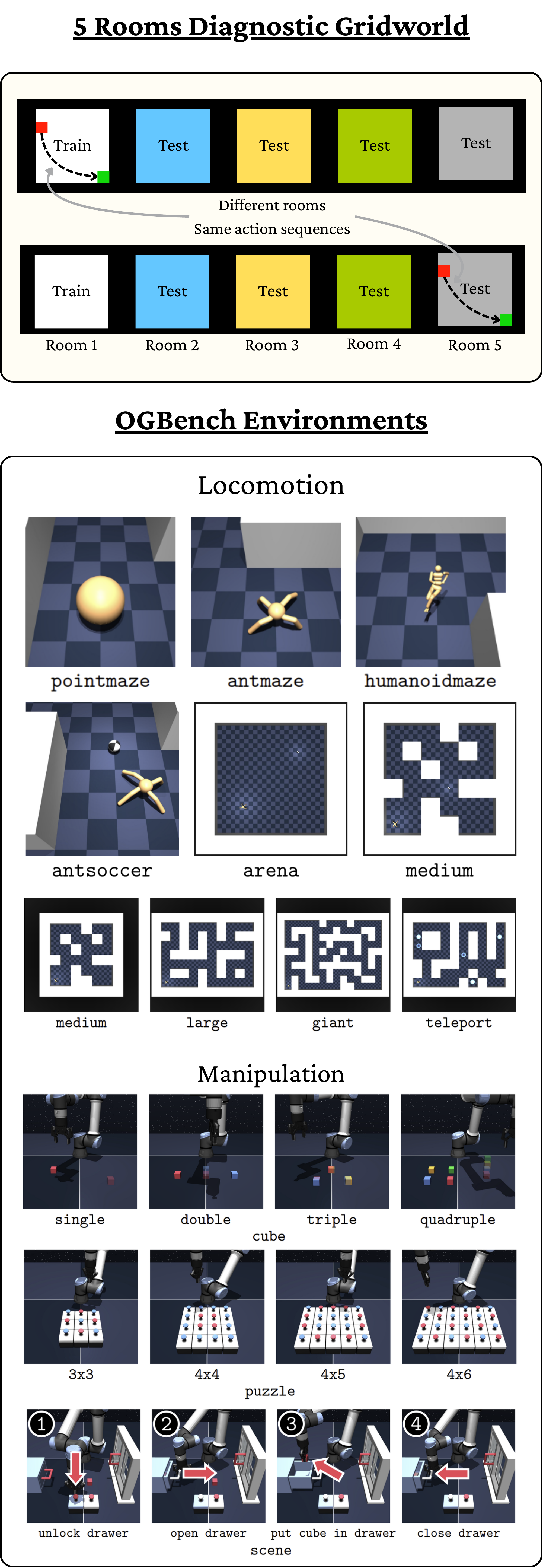}
    \caption{Examples of environments used to benchmark \methodname, including a diagnostic grid world as well as as environments from the OGBench suite.}
    \label{fig:environments}
    \vspace{-31mm}
\end{wrapfigure}

\textbf{(Q1) Skill Reuse.} To test how \methodname enables skill reuse, we design a toy environment consisting of five identical grid-world rooms (figure~\ref{fig:environments}). These rooms differ only in their global $(x, y)$ coordinates, which form the proprioceptive state for the agent. 

We test zero-shot generalization in two settings. First, a goal-conditioned policy trained in a single room must solve the same task in held-out rooms. We compare HIQL to HIQL augmented with a \methodname\ encoder trained on all rooms. Since \methodname\ aliases state-goal pairs reachable by the same action sequences, the policy should transfer through the shared representation. Our experiments confirm this: HIQL+\methodname\ solves all held-out rooms, while HIQL fails to generalize. Second, we scale to 20 identical grid-world rooms, training on 4 and testing on the remaining 16. 

HIQL+\methodname\ generalizes to 12 unseen rooms versus 8 for HIQL alone, further demonstrating that explicitly modeling skill reusability improves transfer.

\subsection{Evaluating Performance Benefits for HRL}

\textbf{(Q2) HRL Integration.} Here, we examine how \methodname benefits HRL by benchmarking on OGBench tasks. We analyze results for HIQL+\methodname and HGCBC+\methodname, as described in Section \ref{sec:method-hrl}.

\subsubsection{Experimental setup}

We evaluate on two task categories from OGBench. The first is the locomotion suite, where various embodiments (point-mass, ant, humanoid) have to navigate throughout a maze. These tasks test learning high-level planning and low-level locomotion skills from offline data. 
We use the navigate datasets for these tasks, collected with a noisy expert policy that wanders the maze by achieving randomly sampled goals. This strategy yields broad coverage with repeating low-level structure, providing opportunities for skill reuse and leveraging \methodname's core capability.

The second category is the manipulation suite, where a 6-DoF UR5e arm performs tasks like stacking cubes, arranging tabletop scenes, or solving puzzles, testing object manipulation and combinatorial generalization. We test on the play datasets, collected by open loop planners with temporally correlated noise. These planners rely on repeated behavior primitives, which we expect \methodname to take advantage of.

These environments feature continuous control and span a range of complexity, from the low-dimensional point maze to the high-dimensional humanoid, allowing us to assess how \methodname scales with state and action space complexity. The robotics domains introduce additional challenges through multi-entity interactions. Together, these environments test \methodname across a diverse range of control difficulties. We defer to the OGBench paper for additional environment and dataset specifications ~\citep{ogbench}.

\subsubsection{OGBench State-Based Performance}

We examine the effect of \methodname on HRL algorithms in Table~\ref{tab:hrl_comparison} by comparing HIQL and HGCBC with their augmented variants in downstream OGBench performance. Notably, HIQL+\methodname improves HIQL by at least 10\% on giant maze variants, 17\% on smaller cube environments, 28\% on smaller puzzle environments, and an impressive 30\% on the scene task. HGCBC+\methodname tends to outperform HGCBC around 10\% for most navigation environments. This provides strong evidence that \methodname's representation benefits HRL in downstream tasks.

\methodname's gains are more limited on antmaze-teleport, antsoccer-medium, and the harder cube and puzzle tasks. The underlying causes remain unclear, but we hypothesize that these environments pose distinct challenges: stochastic dynamics (portals in antmaze-teleport), long horizons with entity-centric generalization (antsoccer-medium, cube), and combinatorial generalization over large configuration spaces (puzzle). Addressing these will likely require advances in the representation structure, such as representations that are invariant to entity permutations or that explicitly support compositional reasoning at the level of the high-level policy.

\begin{table*}[!t]
\centering
\resizebox{\textwidth}{!}{\mergedresults}
\caption{Success rates (\%) on state-based OGBench tasks, reported as mean $\pm$ 95\% CI over 8 seeds. \textbf{(a) HRL Comparison} contrasts HIQL and HGCBC with their \methodname-augmented variants, where we \textbf{bold} values within 95\% of the better method in each pair. \textbf{(b) \methodname{} Ablations} compares variants of \methodname combined with HIQL, where we \textbf{bold} values within 95\% of the best ablation. The final row reports per-algorithm win-rates excluding ties, with the highest \textbf{bolded}.}
\label{tab:main_results}
\vspace{-6mm}
\end{table*}

\begin{wraptable}{r}{0.49\textwidth}
\centering
\resizebox{0.47\textwidth}{!}{
\visualresults
}
\caption{HIQL+CARL vs HIQL on visual environments. Mean success rates (\%) $\pm$ 95\% CI over 4 seeds. Values within 95\% of the best per row are \textbf{bolded}; for win-rates, ties are excluded and the highest is \textbf{bolded}.}
\label{tab:visual_hiql}
\vspace{-10mm}
\end{wraptable}

In robotics domains, \methodname improves HIQL substantially more than it improves HGCBC. We attribute this to the difference in how each algorithm extracts its low-level policy: HIQL uses AWR, which selectively reweights toward high-advantage actions while HGCBC relies on unweighted behavioral cloning. 
Combining our method (to discover where a skill can be reused) with AWR (to discover which skills are optimal) allows us to reuse skills that are more likely to achieve the global task, and thus results in better task performance. While it remains unclear why robotics tasks in particular benefit disproportionately, the pattern suggests a complementary relationship between extracting high-quality skills and knowing where to use them.

\subsubsection{OGBench Pixel-Based Performance}


Visual observations represent a higher-dimensional and more relevant setting for real-world control, making them an important test for \methodname's applicability beyond compact state representations. Thus, we benchmark our HIQL+\methodname variant on OGBench's visual suite without further hyperparameter tuning, and present results in Table~\ref{tab:visual_hiql}. We find significant gains on longer horizon antmaze tasks (over 30\% in large and giant). Robotics domains also see at least a 5\% improvement in scene and larger puzzle tasks, with gains of over 20\% on puzzle-4x4. However, \methodname does not always improve performance. Visual humanoid control is still too difficult even with \methodname's representation, and cube tasks also benefit less than other robotics domains.

\subsection{Ablating \methodname's Design for Capturing Behavioral Similarity}

\textbf{(Q3) Component Ablation.} In this section, we analyze how different parts of our representation learning approach affect its ability to capture behavioral similarity. Our full method optimizes $\mathcal{L}_{\text{InfoNCE}}$ (Eq.~\ref{eq:infonce}) with the action sequence $\mathbf{a_k}$ as the contrastive target. The three ablations modify this along one of two axes. \textit{Single-Action \methodname} replaces $\mathbf{a_k}$ with the first action $a_{k,1}$, keeping the contrastive objective. \textit{Multi-Action Prediction} keeps $\mathbf{a_k}$ but replaces the contrastive loss with direct regression, $\|\xi \big( \phi(s, g_k) \big) - \mathbf{a_k}\|_2^2$, where $\xi$ is a network head which takes in representations of state-goal pairs produced by $\phi$. \textit{Single-Action Prediction} applies both changes.

To compare these ablations, we use UMAP \citep{mcinnes2020umap} to visualize how well each method organizes their latent structure by behavioral similarity. More concretely, we construct state-goal pairs corresponding to motion in cardinal directions for the pointmaze environment from OGBench. We use pointmaze, because the simple dynamics make it easy to reason about the ideal clustering structure: pairs should cluster by the direction of motion they induce. 

\begin{figure*}[!t]
    \centering
    \includegraphics[width=1.0\textwidth]{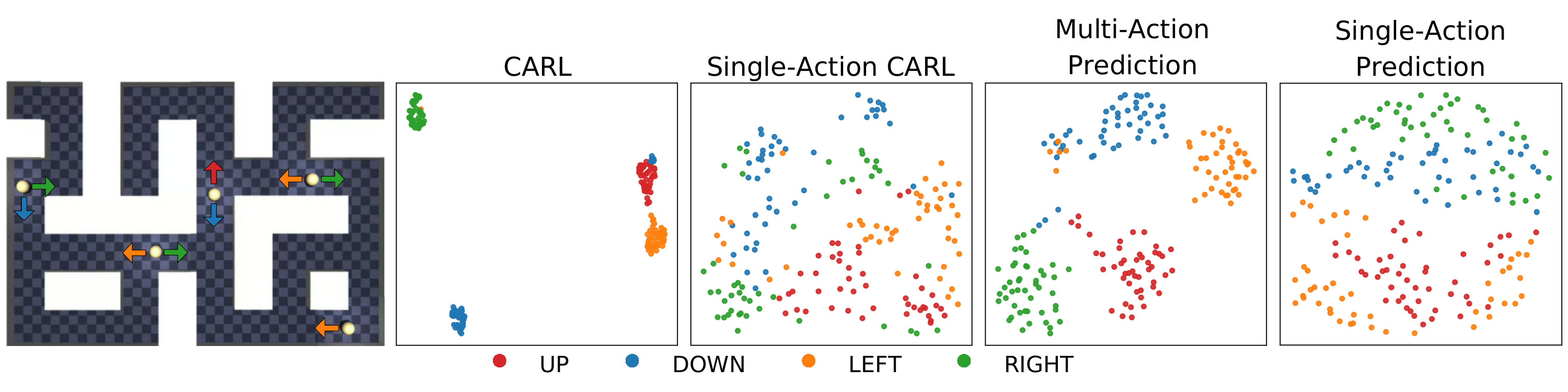}
    \caption{UMAP visualization of learned embeddings colored by behavioral mode. When embeddings are trained using CARL, state-goal pairs that require similar action sequences cluster tightly, even when the raw states differ substantially. Other ablations show more overlap and looser clusters.}
    \vspace{-5mm}
    \label{fig:clusters}
\end{figure*}

Figure~\ref{fig:clusters} shows that only \methodname's combination of action-sequence modeling and a contrastive objective produces tight, disjoint clusters suitable for downstream skill reuse. The ablations yield scattered, overlapping clusters, with the exception of Multi-Action Prediction, whose clusters are looser but still disjoint enough to capture skill reuse meaningfully. This suggests that modeling action sequences is important for capturing behavioral similarity and supporting skill reuse.

We also evaluate downstream performance on state-based OGBench environments in Table~\ref{tab:carl_ablations} to contextualize these differences in representation geometry. \methodname generally outperforms the other variants,  emphasizing the importance of both action-sequence modeling and a contrastive objective. Single-Action \methodname comes closest in performance, while the prediction methods perform comparably to each other. This suggests that the contrastive objective is an effective way to translate behavioral similarity into useful downstream representations.

\begin{figure}[b]
    \centering
    \vspace{-4mm}
    \includegraphics[width=1.0\textwidth]{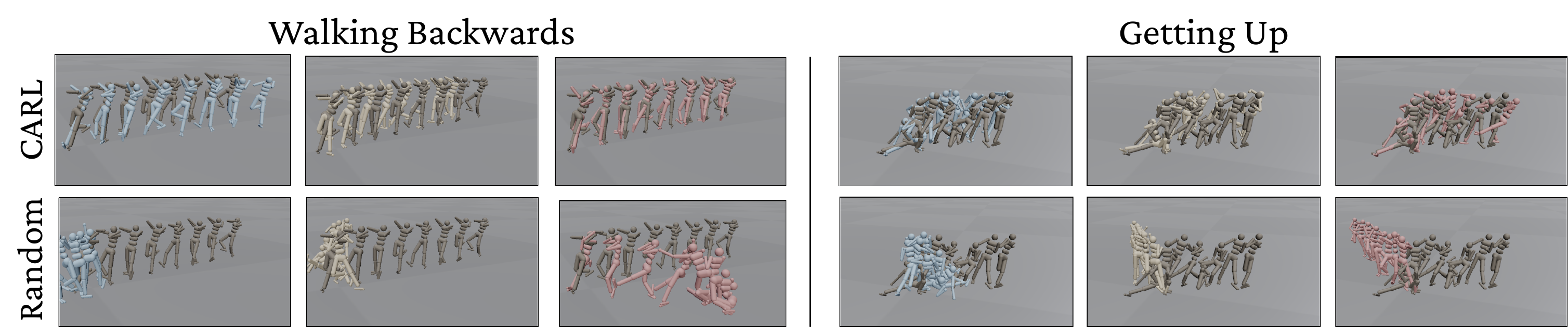}
    \caption{Visualizations of nearest-neighbor trajectories under \methodname and a random encoder for two reference skills (walking backwards and standing up); although these neighbors originate from across the maze, we shift them to share the reference's starting position to highlight their behavioral similarity. The grey trajectory is the reference; blue, yellow, and red are its nearest neighbors.}
    \label{fig:humanoids}
\end{figure}

\subsection{Visualizing \methodname's Representation Structure}

\textbf{(Q4) Skill Structure.} We've already observed how \methodname's geometry tightly clusters state-goal pairs by behavioral similarity. In this section, we examine whether this result extends to higher-dimensional and less structured settings by studying our representations in the humanoid maze environment. 

To probe \methodname's clustering structure, we select nearest neighbors of two reference state-goal pairs, one corresponding to walking backwards and one to standing up, and visualize their rollouts. As shown in figure~\ref{fig:humanoids}, the nearest-neighbor trajectories (blue, yellow, and red) are behaviorally coherent with the gray reference trajectory, whereas the random encoder's neighbors are not. Together with figure~\ref{fig:clusters}, this demonstrates that \methodname's representation organizes less structured, complex skills by their behavioral similarity, even in scaled-up environments.

\section{Limitations and Opportunities for Future Work} 

\subsection{Limitations}

\paragraph{Skill extraction horizon.} In this work, the notion of a skill depends on the choice of extraction horizon $k$. We analyze \methodname's sensitivity to $k$ in Appendix~\ref{app:k-sensitivity} and find that the optimal value is environment-dependent. A more principled approach would remove this dependence, discovering relevant local dynamics structure at various horizons to solve a given task.


\paragraph{Dataset coverage.} Our representation depends on the diversity, coverage, and quality of the offline dataset. Data from a random policy lacks the behaviors needed to learn useful representations, and sparse coverage can cause the collapse of distinct state-goal pairs. Appendix~\ref{app:data-sensitivity} characterizes this sensitivity and shows \methodname does not amplify it relative to HIQL. Addressing these limitations in an online setting will require intelligent data collection or a more robust objective.
\subsection{Future Work}

\paragraph{Skill discovery.} The behavior-centric view suggests a connection between representation learning and skill discovery. Future work could explicitly extract, label, or compose the clusters in \methodname's representation structure, extending prior 
work~\citep{lin2022planning} to yield a unified framework for offline skill discovery and hierarchical control.

\paragraph{Action abstraction.} Our InfoNCE objective aligns state-goal pairs with their short-horizon action sequences, suggesting a duality in how skills can be defined: by the contexts where they are used or by the behaviors that achieve them. Future work could explore this duality by treating action-sequence abstractions, defined by the state-goal pairs they achieve, as skills in their own right. 

\paragraph{High-level policy representations.} \methodname improves the low-level policy, but the high-level policy may be an equally important bottleneck. A natural next step is to develop representations for high-level decision-making: ones that support better planning, guide exploration, or operate over more abstract action primitives.

\paragraph{Online skill discovery.} \methodname currently operates offline, but the behavioral similarity objective could be extended to the online setting. More broadly, addressing the dataset coverage limitations discussed earlier could yield sample-efficiency gains and widen the applicability of hierarchical RL.

\section{Closing Remarks}

We introduced \methodname, a representation learning approach for goal-conditioned offline RL that explicitly captures reusable short-horizon behaviors. Our central claim is that many RL problems contain \emph{local dynamical structure}, lending themselves to a separation between local and global information. We demonstrated that building a notion of local re-usability into the low-level policy of a hierarchy unlocks performance benefits through skill reuse. By learning embeddings that cluster state–goal pairs according to the short-horizon behaviors they admit, our method provides a concrete mechanism for skill reuse that improves existing hierarchical RL frameworks.

\newpage

\bibliography{example_paper}

@InProceedings{impala,
  title = 	 {{IMPALA}: Scalable Distributed Deep-{RL} with Importance Weighted Actor-Learner Architectures},
  author =       {Espeholt, Lasse and Soyer, Hubert and Munos, Remi and Simonyan, Karen and Mnih, Vlad and Ward, Tom and Doron, Yotam and Firoiu, Vlad and Harley, Tim and Dunning, Iain and Legg, Shane and Kavukcuoglu, Koray},
  booktitle = 	 {Proceedings of the 35th International Conference on Machine Learning},
  pages = 	 {1407--1416},
  year = 	 {2018},
  editor = 	 {Dy, Jennifer and Krause, Andreas},
  volume = 	 {80},
  series = 	 {Proceedings of Machine Learning Research},
  month = 	 {10--15 Jul},
  publisher =    {PMLR},
  url = 	 {https://proceedings.mlr.press/v80/espeholt18a.html}
}

@inproceedings{ogbench,
 author = {Park, Seohong and Frans, Kevin and Eysenbach, Benjamin and Levine, Sergey},
 booktitle = {International Conference on Learning Representations},
 editor = {Y. Yue and A. Garg and N. Peng and F. Sha and R. Yu},
 pages = {94937--94982},
 title = {OGBench: Benchmarking Offline Goal-Conditioned RL},
 url = {https://proceedings.iclr.cc/paper_files/paper/2025/file/ecd92623ac899357312aaa8915853699-Paper-Conference.pdf},
 volume = {2025},
 year = {2025}
}

@article{park2024hiql,
  title={Hiql: Offline goal-conditioned rl with latent states as actions},
  author={Park, Seohong and Ghosh, Dibya and Eysenbach, Benjamin and Levine, Sergey},
  journal={Advances in Neural Information Processing Systems},
  volume={36},
  pages={34866--34891},
  year={2023}
}

@article{sutton1999options,
title = {Between MDPs and semi-MDPs: A framework for temporal abstraction in reinforcement learning},
journal = {Artificial Intelligence},
volume = {112},
number = {1},
pages = {181-211},
year = {1999},
issn = {0004-3702},
doi = {https://doi.org/10.1016/S0004-3702(99)00052-1},
url = {https://www.sciencedirect.com/science/article/pii/S0004370299000521},
author = {Richard S. Sutton and Doina Precup and Satinder Singh},
}

@article{dietterich1999maxq,
author = {Dietterich, Thomas G.},
title = {Hierarchical reinforcement learning with the MAXQ value function decomposition},
year = {2000},
issue_date = {August 2000},
publisher = {AI Access Foundation},
address = {El Segundo, CA, USA},
volume = {13},
number = {1},
issn = {1076-9757},
journal = {J. Artif. Int. Res.},
month = nov,
pages = {227–303},
numpages = {77}
}

@inproceedings{
levy2019hac,
title={Hierarchical Reinforcement Learning with Hindsight},
author={Andrew Levy and Robert Platt and Kate Saenko},
booktitle={International Conference on Learning Representations},
year={2019},
url={https://openreview.net/forum?id=ryzECoAcY7},
}

@inproceedings{
agarwal2021behavioralsimilarity,
title={Contrastive Behavioral Similarity Embeddings for Generalization in Reinforcement Learning},
author={Rishabh Agarwal and Marlos C. Machado and Pablo Samuel Castro and Marc G Bellemare},
booktitle={International Conference on Learning Representations},
year={2021},
url={https://openreview.net/forum?id=qda7-sVg84}
}

@inproceedings{
ajay2021opal,
title={{\{}OPAL{\}}: Offline Primitive Discovery for Accelerating Offline Reinforcement Learning},
author={Anurag Ajay and Aviral Kumar and Pulkit Agrawal and Sergey Levine and Ofir Nachum},
booktitle={International Conference on Learning Representations},
year={2021},
url={https://openreview.net/forum?id=V69LGwJ0lIN}
}

@misc{park2025dualgoalrepresentations,
      title={Dual Goal Representations}, 
      author={Seohong Park and Deepinder Mann and Sergey Levine},
      year={2025},
      eprint={2510.06714},
      archivePrefix={arXiv},
      primaryClass={cs.LG},
      url={https://arxiv.org/abs/2510.06714}, 
}

@article{eysenbach2023crl,
  title={Contrastive learning as goal-conditioned reinforcement learning},
  author={Eysenbach, Benjamin and Zhang, Tianjun and Levine, Sergey and Salakhutdinov, Russ R},
  journal={Advances in Neural Information Processing Systems},
  volume={35},
  pages={35603--35620},
  year={2022}
}

@InProceedings{srinivas2020curl,
  title = 	 {{CURL}: Contrastive Unsupervised Representations for Reinforcement Learning},
  author =       {Laskin, Michael and Srinivas, Aravind and Abbeel, Pieter},
  booktitle = 	 {Proceedings of the 37th International Conference on Machine Learning},
  pages = 	 {5639--5650},
  year = 	 {2020},
  editor = 	 {III, Hal Daumé and Singh, Aarti},
  volume = 	 {119},
  series = 	 {Proceedings of Machine Learning Research},
  month = 	 {13--18 Jul},
  publisher =    {PMLR},
  url = 	 {https://proceedings.mlr.press/v119/laskin20a.html}
}

@misc{oord2019cpc,
      title={Representation Learning with Contrastive Predictive Coding}, 
      author={Aaron van den Oord and Yazhe Li and Oriol Vinyals},
      year={2019},
      eprint={1807.03748},
      archivePrefix={arXiv},
      primaryClass={cs.LG},
      url={https://arxiv.org/abs/1807.03748}, 
}

@inproceedings{
zhang2021bisim,
title={Learning Invariant Representations for Reinforcement Learning without Reconstruction},
author={Amy Zhang and Rowan Thomas McAllister and Roberto Calandra and Yarin Gal and Sergey Levine},
booktitle={International Conference on Learning Representations},
year={2021},
url={https://openreview.net/forum?id=-2FCwDKRREu}
}

@inproceedings{hansen2022gcbisim,
  title={Bisimulation makes analogies in goal-conditioned reinforcement learning},
  author={Hansen-Estruch, Philippe and Zhang, Amy and Nair, Ashvin and Yin, Patrick and Levine, Sergey},
  booktitle={International Conference on Machine Learning},
  pages={8407--8426},
  year={2022},
  organization={PMLR}
}

@InProceedings{radford2021clip,
  title = 	 {Learning Transferable Visual Models From Natural Language Supervision},
  author =       {Radford, Alec and Kim, Jong Wook and Hallacy, Chris and Ramesh, Aditya and Goh, Gabriel and Agarwal, Sandhini and Sastry, Girish and Askell, Amanda and Mishkin, Pamela and Clark, Jack and Krueger, Gretchen and Sutskever, Ilya},
  booktitle = 	 {Proceedings of the 38th International Conference on Machine Learning},
  pages = 	 {8748--8763},
  year = 	 {2021},
  editor = 	 {Meila, Marina and Zhang, Tong},
  volume = 	 {139},
  series = 	 {Proceedings of Machine Learning Research},
  month = 	 {18--24 Jul},
  publisher =    {PMLR},
  url = 	 {https://proceedings.mlr.press/v139/radford21a.html}
}

@inproceedings{castro2020scalable,
  title={Scalable methods for computing state similarity in deterministic markov decision processes},
  author={Castro, Pablo Samuel},
  booktitle={Proceedings of the AAAI Conference on Artificial Intelligence},
  volume={34},
  pages={10069--10076},
  year={2020}
}

@inproceedings{islam2023agentcontrollerrepresentationsprincipledoffline,
author = {Islam, Riashat and Tomar, Manan and Lamb, Alex and Efroni, Yonathan and Zang, Hongyu and Didolkar, Aniket and Misra, Dipendra and Li, Xin and Van Seijen, Harm and Des Combes, Remi Tachet and Langford, John},
title = {Principled offline RL in the presence of rich exogenous information},
year = {2023},
publisher = {JMLR.org},
booktitle = {Proceedings of the 40th International Conference on Machine Learning},
articleno = {587},
numpages = {32},
location = {Honolulu, Hawaii, USA},
series = {ICML'23}
}

@article{rudolph2024learningactionbasedrepresentationsusing,
    title={Learning Action-based Representations Using Invariance},
    author={Rudolph, Max and Chuck, Caleb and Black, Kevin and Lvovsky, Misha and Niekum, Scott and Zhang, Amy},
    journal={Reinforcement Learning Journal},
    volume={1},
    pages={342--365},
    year={2024}
}

@article{hafner2022deephierarchicalplanningpixels,
  title={Deep hierarchical planning from pixels},
  author={Hafner, Danijar and Lee, Kuang-Huei and Fischer, Ian and Abbeel, Pieter},
  journal={Advances in Neural Information Processing Systems},
  volume={35},
  pages={26091--26104},
  year={2022}
}

@inproceedings{zhang2021hrllimitations,
 author = {Zhang, Tianren and Guo, Shangqi and Tan, Tian and Hu, Xiaolin and Chen, Feng},
 booktitle = {Advances in Neural Information Processing Systems},
 editor = {H. Larochelle and M. Ranzato and R. Hadsell and M.F. Balcan and H. Lin},
 pages = {21579--21590},
 publisher = {Curran Associates, Inc.},
 title = {Generating Adjacency-Constrained Subgoals in Hierarchical Reinforcement Learning},
 url = {https://proceedings.neurips.cc/paper_files/paper/2020/file/f5f3b8d720f34ebebceb7765e447268b-Paper.pdf},
 volume = {33},
 year = {2020}
}

@inproceedings{
wang2025hierarchical,
title={Hierarchical Reinforcement Learning with Uncertainty-Guided Diffusional Subgoals},
author={Vivienne Huiling Wang and Tinghuai Wang and Joni Pajarinen},
booktitle={Forty-second International Conference on Machine Learning},
year={2025},
url={https://openreview.net/forum?id=1YOYA2zN1j}
}

@inproceedings{dayan1992feudal,
 author = {Dayan, Peter and Hinton, Geoffrey E},
 booktitle = {Advances in Neural Information Processing Systems},
 editor = {S. Hanson and J. Cowan and C. Giles},
 pages = {},
 publisher = {Morgan-Kaufmann},
 title = {Feudal Reinforcement Learning},
 url = {https://proceedings.neurips.cc/paper_files/paper/1992/file/d14220ee66aeec73c49038385428ec4c-Paper.pdf},
 volume = {5},
 year = {1992}
}

@article{eysenbach2019search,
  title={Search on the replay buffer: Bridging planning and reinforcement learning},
  author={Eysenbach, Ben and Salakhutdinov, Russ R and Levine, Sergey},
  journal={Advances in neural information processing systems},
  volume={32},
  year={2019}
}

@misc{mcinnes2020umap,
      title={UMAP: Uniform Manifold Approximation and Projection for Dimension Reduction}, 
      author={Leland McInnes and John Healy and James Melville},
      year={2020},
      eprint={1802.03426},
      archivePrefix={arXiv},
      primaryClass={stat.ML},
      url={https://arxiv.org/abs/1802.03426}, 
}

@article{parr1997reinforcement,
  title={Reinforcement learning with hierarchies of machines},
  author={Parr, Ronald and Russell, Stuart},
  journal={Advances in neural information processing systems},
  volume={10},
  year={1997}
}

@article{nachum2018dataefficienthierarchicalreinforcementlearning,
  title={Data-efficient hierarchical reinforcement learning},
  author={Nachum, Ofir and Gu, Shixiang Shane and Lee, Honglak and Levine, Sergey},
  journal={Advances in neural information processing systems},
  volume={31},
  year={2018}
}

@inproceedings{kingma2017adammethodstochasticoptimization,
  author       = {Diederik P. Kingma and
                  Jimmy Ba},
  editor       = {Yoshua Bengio and
                  Yann LeCun},
  title        = {Adam: {A} Method for Stochastic Optimization},
  booktitle    = {3rd International Conference on Learning Representations, {ICLR} 2015,
                  San Diego, CA, USA, May 7-9, 2015, Conference Track Proceedings},
  year         = {2015},
  url          = {http://arxiv.org/abs/1412.6980},
  bibsource    = {dblp computer science bibliography, https://dblp.org}
}

@misc{hendrycks2023gaussianerrorlinearunits,
      title={Gaussian Error Linear Units (GELUs)}, 
      author={Dan Hendrycks and Kevin Gimpel},
      year={2023},
      eprint={1606.08415},
      archivePrefix={arXiv},
      primaryClass={cs.LG},
      url={https://arxiv.org/abs/1606.08415}, 
}

@inproceedings{
lin2022planning,
title={Planning with Spatial-Temporal Abstraction from Point Clouds for Deformable Object Manipulation},
author={Xingyu Lin and Carl Qi and Yunchu Zhang and Zhiao Huang and Katerina Fragkiadaki and Yunzhu Li and Chuang Gan and David Held},
booktitle={6th Annual Conference on Robot Learning},
year={2022},
url={https://openreview.net/forum?id=tyxyBj2w4vw}
}

@inproceedings{ferns,
author = {Ferns, Norm and Panangaden, Prakash and Precup, Doina},
title = {Metrics for finite Markov decision processes},
year = {2004},
isbn = {0974903906},
publisher = {AUAI Press},
address = {Arlington, Virginia, USA},
booktitle = {Proceedings of the 20th Conference on Uncertainty in Artificial Intelligence},
pages = {162–169},
numpages = {8},
location = {Banff, Canada},
series = {UAI '04}
}

@software{tsujimoto_mjswan,
  author = {Tsujimoto, Tatsuki},
  title  = {{mjswan: MuJoCo Simulation on Web Assembly with Neural Networks}},
  url    = {https://github.com/ttktjmt/mjswan},
  license = {Apache-2.0}
}
\bibliographystyle{icml2026}

\newpage
\appendix
\onecolumn
\section{Training Details}
\label{app:training}

\paragraph{InfoNCE Loss.}  In practice, we calculate our InfoNCE loss using the normalized dot product

\[
f\!\left(\phi(s,g),\,\psi(\mathbf{a}_k)\right)
=
\left\langle
\frac{\phi(s,g)}{\lVert \phi(s,g)\rVert_2},\;
\frac{\psi(\mathbf{a}_k)}{\lVert \psi(\mathbf{a}_k)\rVert_2}
\right\rangle ,
\]

where the loss function is

\[
\mathcal{L}_{\mathrm{InfoNCE}}
=
-\frac{1}{B}
\sum_{i=1}^{B}
\log
\frac{
\exp\!\left( f\!\left(\phi(s^i,g_k^i),\,\psi(\mathbf{a}_k^i)\right) / \tau \right)
}{
\sum_{j=1}^{B}
\exp\!\left( f\!\left(\phi(s^i,g_k^i),\,\psi(\mathbf{a}_k^j)\right) / \tau \right)
}.
\]


Normalizing the InfoNCE score function prevents trivial solutions such as scaling the norms of embeddings, and bounds the scale of logits which makes the temperature parameter meaningful.

\paragraph{Goal Sampling.} 


Given a $k$-step triplet $(s_t, \mathbf{a_k}, s_{t+k})$ sampled from our dataset, we construct positive pairs by uniformly sampling goals from the range $s_{t+1} \hdots s_{t+k}$. In other words, we obtain positive pairs that consist of $(s_t, s_{t+i})$ and $\mathbf{a_k}$, where $1\le i\le k$. This allows us to capture state-goal equivalences for all goals reachable within $k$ steps from a state, aligning closer to our theoretical formulation. In addition, this allows to capture more temporal rich structure in the environment since skills at different horizons may become useful for the high-level policy. The benefits from this strategy are further explored in Appendix~\ref{app:interior}.

\paragraph{Action Sequence Striding.}

For humanoid maze, we found that long action sequences and large action spaces presented computational challenges due to the memory required to store large sequences of large actions. As a result, we introduced an action stride hyperparameter which avoids encoding the entire action sequence. We empirically selected an action stride of 4 for humanoid tasks because we found that this satisfied our computational budget constraints and minimally changed the action sequence. We additionally note that this minimally affects downstream performance, likely because the action sequence with striding captures most of the structure in the action sequence with no striding.

\paragraph{Action Sequence Truncation.}

An edge case arises when constructing state--goal pairs near the end of a trajectory, where the full $k$-step action sequence and goal are unavailable. In this case, we pair the state with the last available goal in the trajectory and pad the action sequence to length $k$ by repeating the final action.

\paragraph{Auxiliary Loss Weighting.} When combining our method with HIQL and HGCBC, we introduce a hyperparameter $\lambda_{aux}$, which weights the policy and value losses against our added objectives, $\mathcal L_{aux}$. Specifically, the final loss is calculated as

\[
\mathcal L_{total} = (1-\lambda_{aux}) (\mathcal L_{value} + \mathcal L_{high} +\mathcal L_{low}) + \lambda_{aux} \mathcal L_{aux}.
\]

Here, $\mathcal L_{low}$ refers to the low-level policy loss, $\mathcal L_{high}$ refers to the high-level policy loss, and $\mathcal{L}_{value}$ refers to the value function loss. $\mathcal L_{aux}$ refers to the auxilary loss applied (which is $\mathcal L_{\text{InfoNCE}}$ for CARL). We report the $\lambda_{aux}$ we select empirically in Table~\ref{tab:hparams}.
\newpage
\section{Implementation Details}

We base our implementation of CARL on OGBench \citep{ogbench}. We run our experiments on an internal GPU Cluster composed of Quadro RTX 5000 GPUs with around 16GB of memory. Similar to times reported in \citet{ogbench}, each run typically takes around 4 hours (state-based tasks) or around 10 hours (pixel-based tasks) when running HIQL+CARL and HGCBC+CARL. We estimate a total of around 10k GPU-hours were required for all experiments, with some additional hours for preliminary experiments.
\label{app:impl}

\begin{table}[th]
\centering
\resizebox{1\linewidth}{!}{
\begin{tabular}{l l}
\toprule
\textbf{Hyperparameter} & \textbf{Value} \\
\midrule
Learning rate & 0.0003 \\
Optimizer & Adam \citep{kingma2017adammethodstochasticoptimization} \\
\# gradient steps & 1000000 (states), 500000 (pixels) \\
Minibatch size & 1024 (states), 256 (pixels) \\
MLP dimensions & (512, 512, 512) \\
Representation architecture (pixel-based) & Impala CNN \citep{impala}\\
Image augmentation probability & 0.5 (pixel-based manipulation), 0 (others) \\
AWR Temperature & 3.0 \\
State-Goal MLP dimensions $\phi(s, g)$ & (512, 512, 512) \\
Action sequence MLP dimensions $\psi(\mathbf{a_k})$ & (256, 256) \\
Nonlinearity & GELU \citep{hendrycks2023gaussianerrorlinearunits} \\
Target smoothing coefficient & 0.005 \\
Discount factor $\gamma$ & 0.995 (\{antmaze, pointmaze\}-giant, humanoidmaze), 0.99 (others) \\
Expectile $\kappa$ & 0.7 \\
Subgoal step & 100 (humanoidmaze), 25 (other locomotion), 10 (others) \\
Representation horizon $k$  & 100 (humanoidmaze), 25 (other locomotion), 10 (others) \\
Action stride & 4 (humanoidmaze), 1 (other) \\
Representation dimension & 100 \\
\singleactionmethodname auxiliary loss weight $\lambda_{\text{aux}}$ & 0.1 (humanoidmaze), 0.5 (other locomotion), 0.1 (others)\\
\methodname auxiliary loss weight $\lambda_{\text{aux}}$ & 0.1 (humanoidmaze), 0.3 (other locomotion), 0.7 (others)\\
Prediction auxiliary loss weight $\lambda_{\text{aux}}$ & 0.1 (humanoidmaze), 0.7 (other locomotion), 0.9 (others) \\
\singleactionmethodname temperature $\tau$ & 0.1 (locomotion), 0.05 (other) \\
\methodname temperature $\tau$ & 0.1 (locomotion),  0.05 (other)\\
Evaluation Episodes & 20 \\

\bottomrule
\end{tabular}
}
\vspace{3mm}
\caption{Hyperparameters used for comparisons across \methodname, and other algorithms used in this paper. "Prediction" refers to parameters used for both Singe-Action and Multi-Action Prediction.}
\label{tab:hparams}
\end{table}

\subsection{Hyperparameters}

We base our implementation of HIQL+\methodname on the implementation of HIQL in the OGBench codebase \citep{ogbench}. To ensure fair comparisons, we keep most hyperparameters the same when integrating \methodname with HIQL and HGCBC. One notable exception was the size of the representation dimension. HIQL and HGCBC originally use representation dimensions of 10, but we change this to 100 in their \methodname variants. For fair comparison, we compare HIQL+\methodname to HIQL with a bigger representation dimension in Table~\ref{tab:ours_vs_hiql_100}. We find that HIQL's performance does not change significantly overall, with effects varying by environment: increasing the dimension causes regressions on pointmaze tasks while substantially improving performance on puzzle-3x3. Therefore, we use the original hyperparameters from \cite{park2024hiql} in our main benchmark.

Our selections of auxiliary loss weights $\lambda_{aux}$ and temperatures $\tau$ were tuned for all algorithms through hyperparameter sweeps of equal effort.

For a full description of the hyperparameters used in our training, refer to Table~\ref{tab:hparams}.

\subsection{Humanoid Visualization Details}

For qualitative humanoid evaluations, we use a custom tool to visualize rollouts of trajectories associated with state-goal pairs in our dataset. We used Muwanx \cite{tsujimoto_mjswan} \footnote{Muwanx Github: \href{https://github.com/ttktjmt/muwanx}{https://github.com/ttktjmt/muwanx}}, a project combining Mujoco, Web Assembly, ThreeJS, and Onnx to bring live Mujoco simulation to web browsers. Building off of this tool, we added various UI features and visualizations suited for our experiments.

To perform the evaluation, we first find all of the nearest neighbors for a reference state-goal pair. We embed every k step segment in our dataset and bin them into sequential groups of 500. From these bins, we select the best matches to the reference state-goal pair, and then select the top 30. Because the state-goal pairs in the humanoidmaze dataset span a variety of global $(x,y)$ locations, we center them to visualize relative differences in joint behavior.

In figure~\ref{fig:humanoids}, we inspect nearest neighbors for different encoders by comparing $k$-step trajectories to a reference behavior. Although it is clear in videos of these 30 matches that representation learning methods cluster better than random encoders, snapshots of trajectories show only part of the picture. Therefore, we choose trajectories representative of the behavior seen in the videos we release on our website: \url{https://sites.google.com/view/behavior-rep/home}.

\section{Additional Experiments}
\label{app:additional_exps}

\subsection{Pre-training vs Co-training in HIQL}
\begin{table}[H]
\centering
\resizebox{0.6\linewidth}{!}{
\begin{tabular}{lcc}
\toprule
Task & Co-train & Pre-train \\
\midrule
\texttt{pointmaze-medium} & \textbf{84.4 $\pm$ 3.1} & 75.0 $\pm$ 6.7 \\
\texttt{pointmaze-large} & \textbf{75.4 $\pm$ 3.1} & 67.0 $\pm$ 6.6 \\
\texttt{pointmaze-giant} & \textbf{64.5 $\pm$ 15.8} & 48.5 $\pm$ 7.6 \\
\texttt{pointmaze-teleport} & \textbf{31.4 $\pm$ 14.0} & 18.0 $\pm$ 5.2 \\
\midrule
\texttt{antmaze-medium} & \textbf{97.9 $\pm$ 0.9} & \textbf{97.2 $\pm$ 2.3} \\
\texttt{antmaze-large} & \textbf{91.9 $\pm$ 2.4} & \textbf{92.8 $\pm$ 2.4} \\
\texttt{antmaze-giant} & \textbf{75.2 $\pm$ 3.8} & 68.2 $\pm$ 5.0 \\
\texttt{antmaze-teleport} & 41.4 $\pm$ 2.1 & \textbf{50.5 $\pm$ 2.0} \\
\midrule
\texttt{humanoidmaze-medium} & \textbf{90.5 $\pm$ 2.4} & 66.8 $\pm$ 4.0 \\
\texttt{humanoidmaze-large} & \textbf{58.3 $\pm$ 3.8} & 17.2 $\pm$ 1.4 \\
\texttt{humanoidmaze-giant} & \textbf{27.2 $\pm$ 3.1} & 4.0 $\pm$ 2.0 \\
\midrule
\texttt{antsoccer-medium} & \textbf{12.8 $\pm$ 2.4} & 10.8 $\pm$ 4.2 \\
\texttt{antsoccer-arena} & \textbf{63.7 $\pm$ 3.1} & 58.5 $\pm$ 4.7 \\
\midrule
\texttt{cube-single} & \textbf{32.8 $\pm$ 3.1} & 20.5 $\pm$ 6.0 \\
\texttt{cube-double} & \textbf{23.4 $\pm$ 4.5} & 8.2 $\pm$ 3.1 \\
\midrule
\texttt{scene} & \textbf{70.5 $\pm$ 5.2} & 49.5 $\pm$ 7.2 \\
\bottomrule
\end{tabular}
}
\vspace{3mm}
\caption{Success rates (\%) comparing Co-train (jointly trained representations) against Pre-train (frozen representations). Results reported as mean $\pm$ 95\% confidence interval; values within 95\% of the best per row are \textbf{bolded}.}
\label{tab:cotrain-vs-pretrain}
\end{table}

We explored two ways of integrating \methodname into HIQL: pretraining, where the subgoal representation is learned with \methodname and then finetuned with low-level policy gradients from HIQL, and co-training, where \methodname's objective is optimized jointly with HIQL's value and policy losses. Pretraining keeps \methodname isolated from HIQL's value learning — the encoder shapes the low-level policy's input but does not interact with the value function. Co-training instead shares the subgoal representation across the low-level policy and value function, fully integrating skill reuse into the algorithm.
Pretraining alone improved performance on most state-based tasks but regressed on humanoid maze tasks. Co-training, by contrast, improved results across all tasks (Table~\ref{tab:cotrain-vs-pretrain}), with the largest gains on longer-horizon tasks. We attribute this to the shared subgoal representation benefiting both the policy and value function, and view it as further evidence for the importance of integrated skill-reuse abstractions in HRL.

\newpage

\subsection{Surface Sampling vs Interior Sampling}
\label{app:interior}
\begin{table}[H]
\centering
\resizebox{0.6\linewidth}{!}{
\begin{tabular}{lcc}
\toprule
Task & Interior & Surface \\
\midrule
\texttt{pointmaze-medium} & \textbf{84.4 $\pm$ 3.1} & 71.4 $\pm$ 5.9 \\
\texttt{pointmaze-large} & \textbf{75.4 $\pm$ 3.1} & 9.9 $\pm$ 9.7 \\
\texttt{pointmaze-giant} & \textbf{64.5 $\pm$ 15.8} & 7.2 $\pm$ 5.9 \\
\texttt{pointmaze-teleport} & \textbf{31.4 $\pm$ 14.0} & 26.7 $\pm$ 6.1 \\
\midrule
\texttt{antmaze-medium} & \textbf{97.9 $\pm$ 0.9} & \textbf{95.0 $\pm$ 4.7} \\
\texttt{antmaze-large} & \textbf{91.9 $\pm$ 2.4} & \textbf{92.4 $\pm$ 3.1} \\
\texttt{antmaze-giant} & \textbf{75.2 $\pm$ 3.8} & \textbf{71.8 $\pm$ 6.4} \\
\texttt{antmaze-teleport} & \textbf{41.4 $\pm$ 2.1} & \textbf{39.7 $\pm$ 5.7} \\
\midrule
\texttt{humanoidmaze-medium} & \textbf{90.5 $\pm$ 2.4} & \textbf{89.3 $\pm$ 3.1} \\
\texttt{humanoidmaze-large} & \textbf{58.3 $\pm$ 3.8} & 52.6 $\pm$ 5.7 \\
\texttt{humanoidmaze-giant} & \textbf{27.2 $\pm$ 3.1} & 21.5 $\pm$ 3.3 \\
\midrule
\texttt{antsoccer-medium} & \textbf{12.8 $\pm$ 2.4} & 11.4 $\pm$ 2.4 \\
\texttt{antsoccer-arena} & \textbf{63.7 $\pm$ 3.1} & 56.9 $\pm$ 4.0 \\
\midrule
\texttt{cube-single} & \textbf{32.8 $\pm$ 3.1} & 28.5 $\pm$ 3.8 \\
\texttt{cube-double} & \textbf{23.4 $\pm$ 4.5} & 15.7 $\pm$ 4.3 \\
\texttt{cube-triple} & \textbf{15.2 $\pm$ 3.5} & 9.9 $\pm$ 4.0 \\
\texttt{cube-quadruple} & 0.1 $\pm$ 0.2 & \textbf{0.3 $\pm$ 0.5} \\
\midrule
\texttt{puzzle-3x3} & \textbf{45.5 $\pm$ 7.3} & \textbf{47.9 $\pm$ 8.0} \\
\texttt{puzzle-4x4} & \textbf{35.2 $\pm$ 7.1} & \textbf{38.2 $\pm$ 5.0} \\
\texttt{puzzle-4x5} & 4.2 $\pm$ 2.4 & \textbf{6.5 $\pm$ 2.1} \\
\texttt{puzzle-4x6} & 3.6 $\pm$ 2.1 & \textbf{4.0 $\pm$ 1.4} \\
\midrule
\texttt{scene} & \textbf{70.5 $\pm$ 5.2} & \textbf{71.6 $\pm$ 3.5} \\
\midrule
\texttt{win-rate} & \textbf{15/22} & 7/22 \\
\bottomrule
\end{tabular}
}
\vspace{3mm}
\caption{Success rates (\%) comparing \methodname\ representations trained with Interior and Surface sampling techniques. Results report mean $\pm$ 95\% confidence interval over 8 seeds. Win-rates for each sampling technique are reported, with the best win-rate bolded.}
\label{tab:interior_vs_surface}
\end{table}

Given a tuple $s_t, \mathbf{a_k}, s_{t+k}$, one simple sampling technique would be to select positive examples as $(s_t, s_{t+k})$ and $\mathbf{a_k}$. We call this surface sampling, because this technique samples state-goal pairs that are on the surface of a $k$-step walk from $s_t$. However, we could also form positive pairs for $(s_t, s_{t+i})$, where $i \leq k$, because these shorter state pairs are still achieved with a portion of these longer action sequences. This also benefits the low-level policy, which may have to reason about state-goal pairs that are smaller than $k$-steps away. We call this interior sampling, because it samples state-goal pairs that are on the interior of the $k$--step walk from $s_t$. Empirically, we find that interior sampling performs better on average, obtaining a win-rate of 15/22 as shown in Table~\ref{tab:interior_vs_surface}.

\newpage

\subsection{Training HIQL with larger representation dimensions}

\begin{table}[H]
\centering
\resizebox{\textwidth}{!}{\repdimresults}
\vspace{2mm}
\caption{Success rates (\%) comparing \methodname-augmented HRL algorithms against their 100 dimensional representation variants and normal variants. Results report mean $\pm$ 95\% confidence interval over 8 seeds, with the performance within 95\% of best \textbf{bolded}. For win-rate calculations, we ignore ties and \textbf{bold} the highest win-rate.}
\label{tab:ours_vs_hiql_100}
\end{table}

We use a different representation dimension than \citet{park2024hiql}, who do not ablate this choice. To verify that our gains do not come from the change in representation dimension alone, Table~\ref{tab:ours_vs_hiql_100} compares our \methodname-augmented HRL algorithms against both their default versions and variants with a 100-dimensional representation. Increasing the representation dimension alone yields mixed results: HIQL (rep\_dim=100) shows degraded performance on most pointmaze environments and improves only on \texttt{puzzle-3x3} and \texttt{cube-single}. This confirms that the increased capacity alone cannot explain our performance gains — \methodname's representation structure is essential for the consistent improvements we see across nearly every task.

\newpage

\subsection{K-step Sensitivity Analysis}
\label{app:k-sensitivity}

\ksweepmedium
\ksweeplarge
\ksweepsmall

We provide sensitivity analysis for our $k$-step hyperparameter in Table~\ref{tab:k_sweep_locomotion} for locomotion environments (except humanoidmaze), Table~\ref{tab:k_sweep_humanoid} for humanoidmaze, and Table~\ref{tab:k_sweep_manipulation} for manipulation environments. For some environments, we observe better performance with larger $k$ (pointmaze-large, pointmaze-giant, and humanoidmaze-large). We also observe the opposite trend where performance drops off after increasing $k$ too much (puzzle-3x3, cube-double, and scene). Other environments are largely robust to the chosen value of the horizon $k$. 

\subsection{Data Sensitivity Analysis}
\label{app:data-sensitivity}

\begin{table}[H]
\centering
\small
\resizebox{1\linewidth}{!}{
\begin{tabular}{llcccc}
\toprule
\textbf{Method} & \textbf{Environment}
& \textbf{0\% Coverage}
& \textbf{25\% Coverage}
& \textbf{50\% Coverage}
& \textbf{75\% Coverage} \\
\midrule
\multirow{3}{*}{CARL + HIQL}
& \texttt{pointmaze-medium} & $10.7 \pm 30.5$ & $76.3 \pm 12.9$ & $80.7 \pm 3.0$ & $\mathbf{85.3 \pm 14.6}$ \\
& \texttt{antmaze-medium} & $11.0 \pm 24.1$ & $\mathbf{95.7 \pm 1.3}$ & $\mathbf{99.7 \pm 1.3}$ & $\mathbf{98.0 \pm 5.2}$ \\
& \texttt{humanoidmaze-medium} & $40.0 \pm 9.0$ & $\mathbf{92.7 \pm 11.2}$ & $\mathbf{88.3 \pm 5.2}$ & $\mathbf{89.3 \pm 1.3}$ \\
\midrule
\multirow{3}{*}{HIQL}
& \texttt{pointmaze-medium} & $12.7 \pm 26.7$ & $62.7 \pm 28.4$ & $\mathbf{72.0 \pm 25.0}$ & $\mathbf{69.7 \pm 12.0}$ \\
& \texttt{antmaze-medium} & $0.0 \pm 0.0$ & $\mathbf{93.0 \pm 6.5}$ & $\mathbf{96.7 \pm 3.9}$ & $\mathbf{95.0 \pm 4.3}$ \\
& \texttt{humanoidmaze-medium} & $33.0 \pm 19.8$ & $\mathbf{89.0 \pm 13.3}$ & $\mathbf{85.0 \pm 9.9}$ & $\mathbf{89.3 \pm 1.3}$ \\
\bottomrule
\end{tabular}
}
\vspace{3mm}
\caption{Robustness to reduced dataset coverage. Success rates (\%) are reported as mean $\pm$ 95\% confidence interval over 3 seeds. Results within 95\% of the best in each row are \textbf{bolded}.}
\label{tab:data_coverage}
\end{table}
\begin{table}[H]
\centering
\small
\resizebox{1\linewidth}{!}{
\begin{tabular}{llcccc}
\toprule
\textbf{Method} & \textbf{Environment}
& \textbf{25\% Removed}
& \textbf{50\% Removed}
& \textbf{75\% Removed}
& \textbf{100\% Removed} \\
\midrule
\multirow{3}{*}{CARL + HIQL}
& \texttt{pointmaze-medium} & $\mathbf{81.0 \pm 1.7}$ & $\mathbf{81.0 \pm 2.6}$ & $59.7 \pm 18.1$ & $19.7 \pm 1.3$ \\
& \texttt{antmaze-medium} & $\mathbf{99.3 \pm 1.3}$ & $\mathbf{96.3 \pm 7.3}$ & $\mathbf{96.7 \pm 3.0}$ & $18.0 \pm 0.0$ \\
& \texttt{humanoidmaze-medium} & $\mathbf{90.3 \pm 7.3}$ & $\mathbf{89.0 \pm 4.3}$ & $\mathbf{86.7 \pm 1.3}$ & $16.0 \pm 5.2$ \\
\midrule
\multirow{3}{*}{HIQL}
& \texttt{pointmaze-medium} & $\mathbf{66.0 \pm 5.2}$ & $\mathbf{63.7 \pm 16.8}$ & $36.3 \pm 11.6$ & $19.0 \pm 4.3$ \\
& \texttt{antmaze-medium} & $\mathbf{96.0 \pm 2.6}$ & $\mathbf{95.3 \pm 8.6}$ & $\mathbf{95.3 \pm 1.3}$ & $15.0 \pm 4.3$ \\
& \texttt{humanoidmaze-medium} & $\mathbf{88.3 \pm 5.2}$ & $\mathbf{87.0 \pm 9.9}$ & $\mathbf{84.0 \pm 6.5}$ & $21.0 \pm 4.3$ \\
\bottomrule
\end{tabular}
}
\vspace{3mm}
\caption{Robustness to dataset imbalance induced by removing action sequences in the left half of the maze that move downwards. Success rates (\%) are reported as mean $\pm$ 95\% confidence interval over 3 seeds. Results within 95\% of the best in each row are \textbf{bolded}.}
\label{tab:data_imbalance}
\end{table}

In this section, we analyze the sensitivity of \methodname to two types of data imbalance. First, to test performance under low coverage, we remove portions of data from the left half of the maze. Specifically, we retain 0, 25, 50, and 75\% of the data present in the left of the maze and test the accuracy of a policy learned with \methodname+HIQL in these conditions and compare it to the performance of HIQL. We report results in Table~\ref{tab:data_coverage} and find that \methodname+HIQL shows a similar drop in performance over time as HIQL. This suggests that \methodname improves the performance of HIQL while making data sensitivity no worse than the baseline. 

Second, we test performance under dataset imbalance. We remove a percentage of action sequences in the left half of the maze that move downwards, biasing the action sequence distribution. We report results for this experiment in Table~\ref{tab:data_imbalance}. Similar to the result in the first case, we find that \methodname+HIQL's drop in performance as we increase dataset imbalance is very similar to the performance drop of HIQL. The similar robustness trends between CARL + HIQL and HIQL suggest that dataset quality impacts performance primarily through the core HRL components rather than the representation. While data diversity and coverage should in principle affect \methodname's performance, these results indicate that \methodname does not further exacerbate existing data sensitivity.

\clearpage

\end{document}